%% file: main.tex
\documentclass[3p,nopreprintline,sort&compress,times]{elsarticle}

\usepackage[utf8]{inputenc}
\usepackage{siunitx}
\usepackage{graphicx}
\graphicspath{{./figs/}}
\usepackage{amsmath,mathtools, cuted}
\usepackage{subcaption}
\usepackage[table,dvipsnames,xcdraw]{xcolor}
\usepackage{booktabs}
\usepackage{multirow}
\usepackage{bm}
\usepackage{textcomp}
\usepackage[most]{tcolorbox}
\usepackage{float}
\usepackage{gensymb}
\usepackage{chemformula}
\usepackage{amsopn}
\usepackage{tabularx}
\usepackage{hyperref}
\hypersetup{
    hidelinks,
    colorlinks=true,
    linkcolor=blue,
    citecolor=blue,
    urlcolor=blue
}
\usepackage{rotating}
\usepackage{setspace} 
\renewcommand{\arraystretch}{0.9} % still useful for non-paragraph cells
\newcolumntype{Y}{>{\raggedright\arraybackslash\setlength{\parskip}{0pt}}X}
\newcolumntype{P}[1]{>{\raggedright\arraybackslash\setlength{\parskip}{0pt}}p{#1}}
\usepackage{overpic}
\usepackage{tikz}
\usepackage{pgfplots}
\pgfplotsset{compat=1.18}  % or 1.17 depending on your Overleaf version
\usepackage{graphicx}
\usepackage{amsmath}

\usepackage{minted}
\usepackage{lineno}
\usepackage{titlesec}

\titleclass{\subsubsubsection}{straight}[\subsubsection]
\newcounter{subsubsubsection}[subsubsection]
\renewcommand\thesubsubsubsection{\thesubsubsection.\arabic{subsubsubsection}}

\titleformat{\subsubsubsection}
  {\normalfont\normalsize\emph}{\thesubsubsubsection}{1em}{}

\titlespacing*{\subsubsubsection}{0pt}{3.25ex plus 1ex minus .2ex}{0ex plus .2ex}

\usepackage{xr-hyper}
\usepackage{cleveref}

\begin{document}

\sloppy

\begin{frontmatter}

\title{DataScribe: An AI-Native, Policy-Aligned Web Platform for Multi-Objective Materials Design and Discovery}

\author[1]{Divyanshu Singh}
\author[2]{Do\u{g}uhan Sar\i{}t\"urk}
\author[1]{Cameron Le}
\author[2]{Md Shafiqul Islam}
\author[2]{Raymundo Arroyave}
\author[2]{Vahid Attari\corref{cor1}}
\cortext[cor1]{Corresponding author}
\ead{attari.v@tamu.edu}

\affiliation[1]{organization={Department of Computer Science, Texas A\&M University},
                city={College Station},
                state={TX},
                postcode={77843},
                country={USA}}

\affiliation[2]{organization={Department of Materials Science and Engineering, Texas A\&M University},
                city={College Station},
                state={TX},
                postcode={77843},
                country={USA}}

\begin{abstract}

The acceleration of materials discovery requires digital platforms that go beyond data repositories to embed learning, optimization, and decision-making directly into research workflows. We introduce \emph{DataScribe}, an AI-native, cloud-based materials discovery platform that unifies heterogeneous experimental and computational data through ontology-backed ingestion and machine-actionable knowledge graphs. The platform integrates FAIR-compliant metadata capture, schema and unit harmonization, uncertainty-aware surrogate modeling, and native multi-objective multi-fidelity Bayesian optimization, enabling closed-loop propose–measure–learn workflows across experimental and computational pipelines. DataScribe functions as an application-layer intelligence stack, coupling data governance, optimization, and explainability rather than treating them as downstream add-ons. We validate the platform through case studies in electrochemical materials and high-entropy alloys, demonstrating end-to-end data fusion, real-time optimization, and reproducible exploration of multi-objective trade spaces. By embedding optimization engines, machine learning, and unified access to public and private scientific data directly within the data infrastructure, and by supporting open, free use for academic and non-profit researchers, DataScribe functions as a general-purpose application-layer backbone for laboratories of any scale, including self-driving laboratories and geographically distributed materials acceleration platforms, with built-in support for performance, sustainability, and supply-chain–aware objectives.

\end{abstract}


%\begin{graphicalabstract}
%\includegraphics{grabs}
%\end{graphicalabstract}

%\begin{highlights}
%\item Research highlight 1
%\item Research highlight 2
%\end{highlights}

\end{frontmatter}

%\tableofcontents % remove when submissions

\input{01_introduction}

\input{02_0_results-and-discussion}

\input{02_1_datascribe-architecture-and-integrations}

\input{02_2_heterogeneous-data-unification-workflow}
\input{02_4_datascribe-data-management-interface}
\input{07_apis}
\input{08_datascribe-agents}

\input{09_datascribe-application-interface}

\input{10_conclusion}
\input{11_methods}
\input{12_codeAvailability}

\bibliographystyle{elsarticle-num}
\bibliography{refs,refs_vahid,refs_vahid_new}

\end{document}

%% file: 01_introduction.tex
\section{Introduction}

%	•	Problem Context
%	-- 	Materials discovery cycle: >10 years from concept to commercialization.
%	-	Limitations in current computational & experimental integration tools.
%	-	Data fragmentation across labs, formats, and repositories.
%	•	Opportunity
%	-	Digital platforms for materials R&D are emerging but often siloed.
%	-	Need for interoperable, AI-native platforms that combine data ingestion, ML model training, and active learning.
%	•	Positioning DataScribe
%	•	Bridges experiment–computation gap.
%	•	Facilitates reproducible research.
%	•	Optimized for US market needs: energy security, defense materials, electronic materials.

Modern challenges across science, technology, and society increasingly demand rapid materials innovation, motivating growing interest in materials acceleration frameworks. Materials underpin semiconductors, aerospace, energy, and infrastructure systems, making them central to technological capability across a wide range of applications. Yet the materials development cycle still routinely extends over a decade from initial concept to deployment, even in systems where the underlying physical and chemical mechanisms are well understood. This delay reflects not only scientific complexity but also structural limitations in current research workflows, including fragmented datasets, incompatible data formats, and weakly integrated experimental and computational activities. Addressing these limitations requires platforms that support closed-loop coupling between data generation, modeling, and synthesis, enabling systematic learning across length and time scales in materials development \cite{nurkincandle,lockhorst2025agenda,sapkota2025ai,zhang2025evolving}.

The A–Z framework summarized in Table~\ref{tab:SDL_AZ_Framework} delineates the set of foundational tasks and system-level integrations required to enable AI-native, closed-loop discovery workflows in practice. Spanning \textbf{A (Architecture Definition)} through \textbf{Z (Zero-Error Learning)}, each element represents a distinct but interdependent capability necessary for the construction and operation of materials acceleration platforms (MAPs) that integrate automation, data infrastructure, and active learning across workflows \cite{flores2020materials,szymanski2023autonomous}. By explicitly incorporating experiment design, workflow orchestration, optimization, and governance within a unified digital architecture, the framework supports iterative coupling between hypothesis generation, synthesis, and validation over successive cycles. In this context, the framework provides a basis for interoperable Self Driving Labs (SDLs) \cite{tabor2018accelerating,yoshikawa2025self,hase2019next,noh2019inverse,abolhasani2023rise} operating under shared architectural and data principles across multiple implementations. Within this structure, Bayesian optimization acts as the mechanism linking learning (L), experiment selection (F), optimization (O), and physical constraints (P), enabling sequential decision-making informed by accumulated data and prior outcomes. For initial demonstrations, simplified materials systems, such as mixtures with straightforward actuation and directly measurable properties provide a tractable setting for instantiating SDL concepts without the complexity of full materials synthesis.

\begin{table}[ht!]
\centering
\caption{\textbf{A--Z Framework for Building and Operating a Self-Driving Laboratory (SDL)}}
\renewcommand{\arraystretch}{1.15}
\scriptsize
\begin{tabular}{p{0.04\textwidth} p{0.21\textwidth} p{0.65\textwidth}}
\toprule
\textbf{Letter} & \textbf{Task} & \textbf{Description} \\
\midrule
A & Architecture Definition & Scientific goal, problem scope, and control hierarchy (Establishing hypotheses and constraints). \\
B & Benchmarking & Establish baselines using literature or initial manual experiments to define performance targets. \\
C & Control Framework & Design automation layers—scheduler, device controller, data acquisition, and safety interlocks. \\
D & Data Infrastructure & Build FAIR-compliant data storage, metadata schema, and access protocols for (un)structured data. \\
E & Experiment Design (DoE) & Implement initial sampling (e.g., Latin Hypercube, Sobol) to bootstrap model training. \\
F & Feedback Loop (Active Learning) & Couple real-time data acquisition to Bayesian or reinforcement learning models for adaptive experimentation. \\
G & Generative Model Synergy & Use VAE, GFlowNet, or LLM-based generators to propose new compositions or reaction conditions. \\
H & Hardware Integration & Integrate robotic systems, reactors, or synthesis modules with calibration and redundancy. \\
I & Instrument Interfacing & Standardize communication across instruments (TCP/IP, serial, OPC-UA) with robust driver management. \\
J & Job Scheduler & Orchestrate concurrent experiments and compute tasks (e.g., via Kubernetes, Airflow, or Prefect). \\
K & Knowledge Graph & Build ontologies linking materials, processes, and properties and integrate with APIs (e.g., OPTIMADE). \\
L & Learning Pipeline & Define preprocessing, model training, validation, and uncertainty quantification procedures. \\
M & Measurement Automation & Automate data collection from sensors, balances, spectrometers, and ensure continuous calibration. \\
N & Networking and Cloud & Ensure secure connectivity between edge devices, local compute, and cloud storage. \\
O & Optimization Engine & Implement Bayesian or multi-objective optimization (BoTorch, Optuna) with constraint handling. \\
P & Process Modeling & Couple physical models or digital twins to guide AI search within feasible parameter space. \\
Q & Quality Assurance & Monitor anomalies, sensor drift, and failed experiments with automated validation protocols. \\
R & Resource Management & Dynamically allocate compute, reagents, and instruments based on experiment priority. \\
S & Safety and Compliance & Enforce EHS compliance, automated shutdowns, and traceable chemical inventory management. \\
T & Traceability & Assign UUIDs to every experiment linking recipes, conditions, and outputs for reproducibility. \\
U & User Interface & Develop intuitive dashboards for control, visualization, and performance tracking. \\
V & Version Control & Apply GitOps for all models, scripts, and datasets; ensure reproducibility via environment snapshots. \\
W & Workflow Orchestration & Automate end-to-end loop: design → synthesis → characterization → analysis → learning update. \\
X & eXplainability & Use SHAP/LIME or attention-based visualization to interpret AI-driven experiment selection. \\
Y & Yield and Performance Metrics & Define KPIs such as discovery velocity, throughput, and cost per experiment cycle. \\
Z & Zero-Error Learning & Enable continuous calibration and self-improvement toward minimal error and human oversight. \\
\bottomrule
\end{tabular}
\label{tab:SDL_AZ_Framework}
\end{table}

Realizing such closed-loop discovery workflows requires a robust digital backbone capable of unifying A--Z tasks across data, models, and workflows spanning experimental and computational environments. Platforms such as \textbf{DataScribe.cloud} are designed to serve this role by supporting FAIR data principles~\cite{wilkinson2016fair,mons2018data}, harmonizing metadata standards~\cite{kush2020fair,marketakis2013harmonizing,diaz2025data}, and providing ontology-driven interfaces~\cite{franconi2011quelo,ryabinin2019ontology,christophides1999ontology} that enable interoperability across heterogeneous laboratory settings~\cite{sadeghi2024interoperability}. By coupling user-generated data with structured repositories, adaptive Application Programming Interfaces (APIs), and graph-based data models, DataScribe.cloud links synthesis, characterization, and modeling data streams into a unified, queryable environment. This shared infrastructure supports the deployment of Bayesian optimization engines and generative models across distributed self-driving laboratory workflows, enabling cross-platform learning, reproducibility, and continuous accumulation of materials knowledge. In this way, digitally integrated data infrastructures provide a mechanism for coordinating automation efforts across otherwise disconnected discovery pipelines.

\begin{figure}
    \centering
    \includegraphics[width=0.5\linewidth]{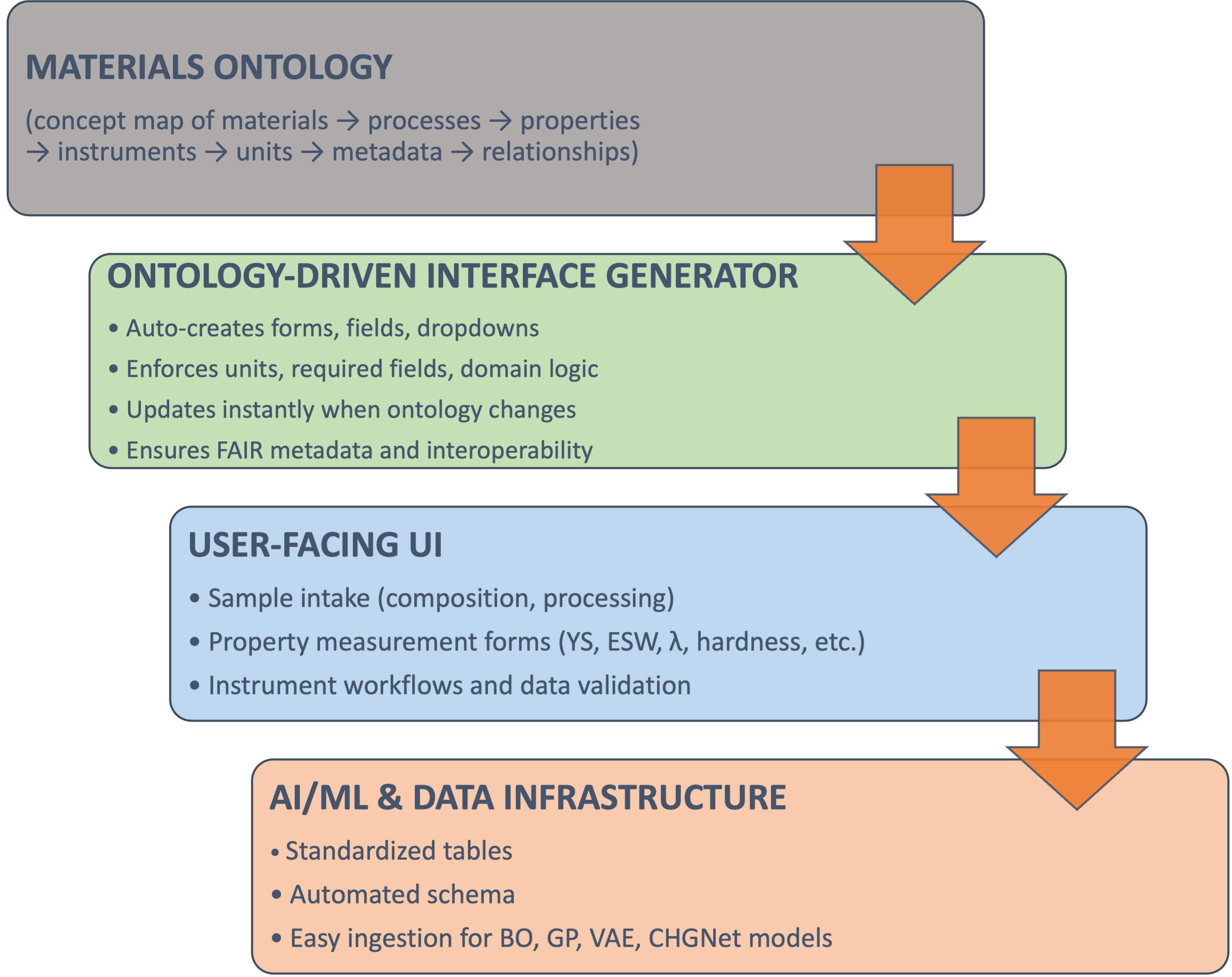}
    \caption{Ontology-Driven Interfaces as a Self-Evolving Layer for Materials Data Platforms. Four-tier architecture showing how materials ontology (top) automatically generates data capture interfaces (second layer) that enforce FAIR compliance and semantic consistency, populate user-facing forms for sample intake and property measurements (third layer), and produce standardized, ML-ready data tables (bottom). Orange arrows indicate automated propagation of domain knowledge through system layers, enabling seamless integration of Bayesian optimization (BO), Gaussian processes (GP), variational autoencoders (VAE), and generative models (CHGNet) without manual schema engineering.}
    \label{fig:Ontology-driven interface}
\end{figure}

Government-led efforts have long emphasized the importance of coordinated materials infrastructures. In the United States, the ICME cyberinfrastructure initiative~\cite{allison2006integrated} articulated and advanced the integration of experimental datasets, multiscale simulations, and process--structure--property workflows under the leadership of the National Academy of Engineering and the National Science Foundation~\cite{arnold2014robust}. In the United Kingdom, the National Physical Laboratory (NPL) has aligned advanced materials metrology with national priorities spanning net-zero, energy, health, and security~\cite{pollard2024advanced,gilmore2022metrology}. At the international level, the Versailles Project on Advanced Materials and Standards (VAMAS), established under G7 auspices, continues to coordinate interlaboratory protocols across domains ranging from wear testing to nanostructure measurements~\cite{minelli2022versailles}. 

More recently, large-scale computing infrastructures such as China’s CNGrid have integrated high-performance computing, data grids, and workflow engines under the 863 Hi-Tech Program~\cite{depei2004cngrid}, supported by a distributed network of national supercomputing centers hosting systems including Tianhe-2~\cite{lu2020design,zhang2014optimizing}. In Europe, the e-Science programme~\cite{hey2003science} and the Worldwide LHC Computing Grid (WLCG)~\cite{shiers2007worldwide} similarly exemplify distributed cyberinfrastructure enabling domain-scale science through international collaboration. Collectively, these initiatives have established foundational standards and benchmarking capabilities, but they remain primarily focused on data curation, protocol harmonization, and computational access, rather than on embedding optimization, active learning, or adaptive decision-making directly into materials discovery workflows.

%% national-scale digital infrastructure 
Building on these efforts, the U.S. National Institute of Standards and Technology (NIST) has advanced metadata schemas, ontologies, and FAIR-compliant frameworks through its Office of Data and Informatics. The NIST--JARVIS platform complements this work by integrating density functional theory (DFT), force-field development, machine learning, and benchmarking into a widely adopted computational resource~\cite{choudhary2020joint}. Together, these initiatives serve as reference implementations for computational--experimental integration, yet they continue to function primarily as repositories and toolkits rather than as application-layer systems capable of guiding design decisions in real time. In the United States, these efforts are supported by a broader cyberinfrastructure backbone that provides the computational and data-transfer capacity upon which domain-specific platforms are constructed. The National Science Foundation’s progression from the TeraGrid~\cite{mervis2001nsf} to XSEDE~\cite{towns2014xsede} and more recently ACCESS~\cite{boerner2023access} has unified national supercomputing resources, while Globus~\cite{foster2011globus} has emerged as a standard for secure, high-throughput data transfer and the Open Science Grid~\cite{ruth2007open} has established a distributed computing environment at national scale. Web-based science gateways such as nanoHUB~\cite{klimeck2008nanohub} further illustrate the role of shared platforms in broadening access to computational tools. Collectively, these infrastructures provide scale, reliability, and partial accessibility, but they do not, on their own, deliver domain-aware orchestration across heterogeneous experimental and computational workflows required for materials acceleration platforms.

%% domain-specific computational and data ecosystems 
Beyond national-scale digital infrastructure, a complementary set of domain-specific computational and data ecosystems has become increasingly important. Platforms such as the Materials Project~\cite{jain2013commentary} and the Open Quantum Materials Database (OQMD) have established large-scale repositories of density functional theory (DFT) calculations~\cite{andronico2011infrastructures}, while AFLOW has generated millions of high-throughput entries spanning thermodynamic and mechanical properties~\cite{curtarolo2012aflow}. Similarly, NOMAD has aggregated materials data from diverse sources under a common metadata schema to support FAIR principles~\cite{shiers2007worldwide}, and Japan’s Materials Integration by Network Technology (MInt) initiative~\cite{demura2024challenges} has coupled multiscale simulations with experimental data for structural materials. Collectively, these platforms provide extensive datasets, increasingly interoperable schemas, and explicit links between modeling and manufacturing. However, most remain focused on specific material classes, property domains, or data modalities, limiting their applicability as general-purpose infrastructures for integrated materials discovery.

\input{table2}
These initiatives collectively reflect substantial momentum toward centralized, computationally accessible materials infrastructures. However, they largely remain partitioned, addressing standards development, large-scale computation, or data aggregation in isolation, without providing the autonomous, cross-modal integration required for sustained acceleration. A persistent gap is the absence of an AI-native application layer capable of unifying heterogeneous data sources, encoding metadata within interoperable ontologies, and coupling these resources directly with optimization and decision-making. The present manuscript introduces \emph{DataScribe}, a web-native platform designed to address this gap by: (i) unifying disparate data sources through ontology-backed ingestion and FAIR-compliant metadata; (ii) embedding multi-objective Bayesian optimization with uncertainty-aware and physics-informed sampling; (iii) integrating laboratory and high-performance computing workflows within closed-loop propose--measure--learn cycles; and (iv) supporting design decisions informed by constraints related to critical materials, embodied carbon, and supply-chain considerations. These capabilities position \emph{DataScribe} as an application-layer materials acceleration platform that enables integrated, data-driven discovery workflows across experimental and computational environments.

%% file: table2.tex
%\begin{sidewaystable}
\begin{table}[!ht]
\scriptsize
\caption{\textbf{High-level comparison of major materials informatics and national cyberinfrastructure platforms.}}
\label{tab:platform_comparison}
\renewcommand{\arraystretch}{1.3}
\begin{tabularx}{\linewidth}{X X p{3.1cm} X X X}
\toprule
\textbf{Platform} & \textbf{Host / Origin} & \textbf{Primary Focus} & \textbf{Data Scale} & \textbf{AI / Optimization} & \textbf{FAIR / Metadata} \\
\midrule

\multicolumn{6}{l}{\textbf{National Standards \& Initiatives}} \\
%\midrule
ICME Cyberinfra. & NSF / NAE (USA) & ICME workflows & Federated datasets & Workflow-level only & ICME metadata \\
NPL (UK) & NPL (UK) & Materials metrology & Reference datasets & Standards-driven & metrology schemas \\
VAMAS & G7 Consortium & Interlaboratory standards & Round-robin data & None & Global testing standards \\
NIST ODI & NIST (USA) & FAIR schemas; registries & Curated datasets & Limited tools & FAIR-by-design \\
NIST-JARVIS & NIST (USA) & DFT + ML benchmarks & $>$80k materials & ML models, GNNs & FAIR-compliant \\
\midrule

\multicolumn{6}{l}{\textbf{Cyberinfrastructure Backbones}} \\
%\midrule
ACCESS / XSEDE & NSF (USA) & National HPC backbone & Multi-center HPC & None & N/A \\
Globus & Argonne / UChicago & Secure data movement & HPC + cloud workflows & Workflow automation & FAIR-aligned transfers \\
OSG & DOE / NSF (USA) & Distributed HTC grid & 80+ institutions & None & Discipline standards \\
nanoHUB & Purdue / NSF & Web simulation tools & 100+ tools & Limited AI & Partial metadata \\
WLCG & CERN / Intl. & Distributed HEP computing & 170+ centers & Workflow-level & HEP data standards \\
\midrule

\multicolumn{6}{l}{\textbf{Domain-Specific Platforms}} \\
%\midrule
Materials Project & LBNL (USA) & DFT database + APIs & $>$150k entries & API-based ML & DOE FAIR schemas \\
AFLOW & Duke Univ. & Automated HT DFT & $>$3.5M entries & ML datasets & FAIR-compliant \\
NOMAD & EU Center & Repository + metadata hub & Millions of calcs & AI analytics & Full FAIR \\
MInt & NIMS (Japan) & Multiscale alloys workflows & Large industrial data & Simulation coupling & Metadata-enabled \\
\midrule

\textbf{DataScribe} & Texas A\&M (USA) & Unified ingestion + knowledge graphs + MOBO & Heterogeneous exp/comp datasets & \textbf{Real-time MOBO + active learning} & \textbf{Ontology-driven FAIR compliance} \\
\bottomrule
\end{tabularx}
\end{table}

%% file: 02_0_results-and-discussion.tex
\section{Results and Discussion}
\subsection{Conceptual Overview of DataScribe Platform}

%Despite decades of investment in cyberinfrastructure (e.g., ACCESS/XSEDE, Globus, OSG) and domain platforms (e.g., NIST/ODI, JARVIS, Materials Project, NOMAD, AFLOW), no existing system provides a unified application layer that couples heterogeneous experimental and computational data with optimization, uncertainty quantification, and policy-driven objectives. Current resources either deliver raw compute and data movement, or serve as repositories and workflow toolkits. Initiatives such as AiiDA, Citrine, and Globus Flows have begun addressing aspects of workflow orchestration, but none operationalize an ontology-backed intelligence stack that closes the loop between data, simulation, and decision-making. 

\begin{figure}[!ht]
    \centering
    \includegraphics[width=0.98\linewidth]{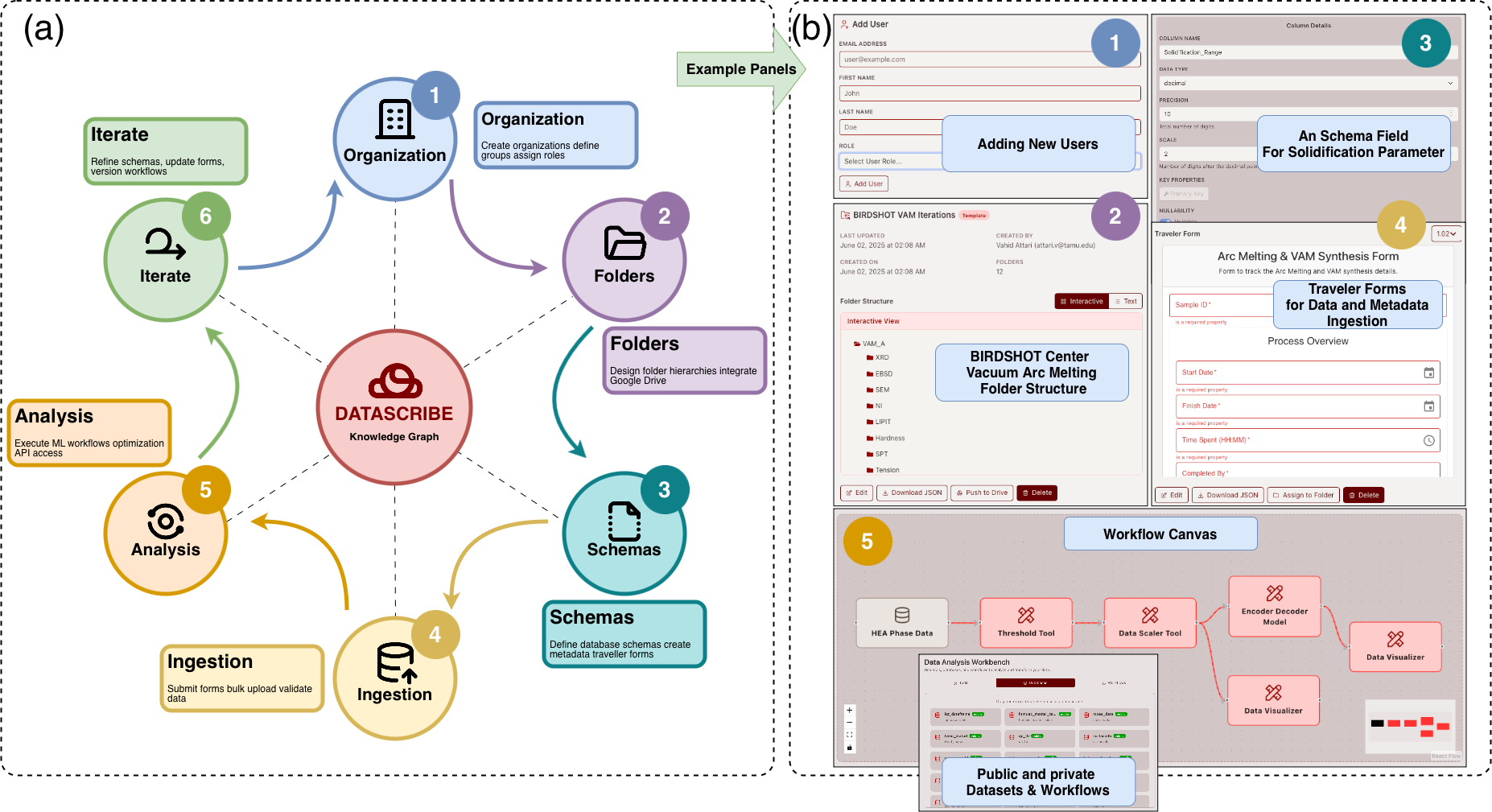}
    \caption{(a) Overview of the DataScribe user workflow spanning six stages: organization and access management, project and folder architecture, schema and traveler form design, data ingestion and population, data analysis and workflow execution, and iteration and collaboration. These stages are realized through five platform interfaces, with the outer iteration loop representing continuous refinement as research programs evolve. (b) Example interface panels illustrating key functionalities: user onboarding, hierarchical experiment and characterization folders (shown here for the BIRDSHOT vacuum arc-melting campaign), schema fields for solidification-relevant parameters, traveler forms for structured data and metadata capture, and a workflow canvas integrating public and private datasets with modular AI tools (e.g., thresholding, scaling, encoder–decoder models, and visualizers).}
    \label{fig:workflow-overview}
\end{figure}

\emph{DataScribe} is an AI-native application-layer platform in which core logic, orchestration, and decision-making are mediated through machine-learning models rather than fixed, rule-based pipelines. The platform embeds model-centric reasoning within data ingestion, schema definition, analysis workflows, and optimization, enabling adaptive decision-making across heterogeneous materials datasets. Within materials discovery programs, this model-first architecture operates as an intermediary between data sources and automated experimental or computational environments, supporting more direct transitions from raw information to actionable design decisions. Figure~\ref{fig:workflow-overview} illustrates how this application-layer functionality is instantiated through structured workflows linking governance, schema creation, ingestion, and analysis.

In this architecture, learning, inference, and uncertainty modeling are treated as integral components of workflow execution rather than downstream analysis steps. Instead of invoking models after data aggregation, \emph{DataScribe} applies them throughout the application layer to classify, interpret, prioritize, and optimize data as it enters and propagates through the system. This approach supports continuous hypothesis updating, dynamic allocation of computational or experimental resources, and closed-loop coordination of discovery workflows. Ontology-backed ingestion ensures that heterogeneous data, including experimental, computational, and literature-derived sources, enter the platform in a semantically aligned and machine-interpretable form. By enforcing structure, units, semantic meaning, and provenance at ingestion, \emph{DataScribe} enables multimodal integration of data types such as microscopy, CALPHAD outputs, cyclic voltammetry (CV), galvanostatic charge--discharge (GCD) curves, and phase-field simulations within a coherent analytical framework.

Embedding multi-objective Bayesian optimization directly into the ingestion and analysis pipeline allows acquisition functions, Pareto fronts, and uncertainty estimates to be updated as new data become available, reducing latency between measurement, analysis, and decision-making. For high-dimensional materials problems, including high-entropy alloys, solidification pathways, and electrochemical materials systems, this tightly coupled optimization framework supports near–real-time steering of propose--measure--learn cycles and concentrates sampling in regions of highest expected value within the design space.

%%%%%
%%%%%
%%%%%

Building on this conceptual framework, \emph{DataScribe} introduces a suite of integrated capabilities that operationalize intelligence at the application layer:

\begin{enumerate}
\item \textbf{Hierarchical collaboration and governance.} Secure multi-organization workspaces with role-based permissions, selective data sharing, and administrative controls for onboarding, lifecycle management, and compliance (Fig.~\ref{Sfig:web-frontend-screenshots}(a)).
\item \textbf{Ontology-backed heterogeneous data ingestion.} Ingests experimental, computational, and literature-derived data—including tabular properties, electrochemical curves, microscopy/spectroscopy, CALPHAD, phase-field, and DFT outputs—into a unified knowledge graph with provenance tracking, FAIR metadata, and semantic search.
\item \textbf{Embedded real-time MOBO with uncertainty.} Provides native multi-objective optimization with Gaussian-process and surrogate backends, flexible objectives, and calibrated uncertainty, enabling performance-, cost-, carbon-, and supply-risk-aware design.
\item \textbf{Closed-loop orchestration across lab and HPC.} Connectors to laboratory instruments and HPC/cloud schedulers automate propose–measure–learn cycles, reducing iteration times in automated or semi-automated environments.
\item \textbf{Hybrid digital twins combining physics and ML.} Integrates CALPHAD, phase-field, and DFT models with learned surrogates to support physics-informed extrapolation in sparse, multi-objective regimes.
\item \textbf{Policy-aware optimization.} Incorporates critical-mineral risk, domestic sourcing likelihood, cost-of-risk, and CO$_2$ budgets directly into design loops, enabling strategic alignment with industrial and national priorities.
\item \textbf{Interoperability with global cyberinfrastructure.} Integrates with ACCESS/HPC, Globus endpoints, OSG/WLCG-style compute pools, and international research grids to support reproducible and portable discovery campaigns.
\item \textbf{Human-in-the-loop explainability and governance.} Provides interactive Pareto fronts, attribution tools, and uncertainty overlays alongside full provenance and licensing metadata for expert oversight.
\item \textbf{Agentic chatbot integration.} Embeds a domain-tuned conversational agent capable of querying datasets, interpreting optimization trajectories, and contextualizing results using literature and historical experiments.
\end{enumerate}

These capabilities enable the consolidation of otherwise fragmented materials workflows into a unified, AI-native discovery layer that supports faster iteration, systematic knowledge retention, and more informed materials decision-making. In applied settings, this integration shortens development cycles and reduces uncertainty in materials selection and qualification, while in research contexts it facilitates interoperable, reusable knowledge across tools, institutions, and workflows. At the system level, such platforms provide a foundation for aligning materials development with broader constraints related to energy, sustainability, and manufacturing readiness.

%%%%
%%%%%
%%%%
%%%%

\section{Scientific Results Enabled by DataScribe}

The DataScribe platform includes a visual, node-based \emph{Workflow Management System} (\textbf{WfMS}) that enables users to construct, debug, and execute complex data-processing and machine-learning workflows without requiring direct code development. The WfMS is tightly integrated with the platform’s \emph{Electronic Laboratory Notebook} (\textbf{ELN}), allowing experimental data, associated metadata, and instrument outputs to propagate directly into workflow nodes for downstream analysis and optimization. In addition to user-defined workflows, the platform provides a set of predefined workflows that can be executed either through the WfMS interface (see Fig.~\ref{fig:workflow-overview}(b)) or via reactive, declarative, widget-based graphical user interfaces (see Fig.~\ref{fig:bayesian-mobo-workflow}, Fig.~\ref{Sfig:redox-predict-workflow}, Fig.~\ref{Sfig:single-obj-bo-workflow}, and Fig.~\ref{Sfig:encoder-decoder-workflow}).

At the time of writing, four major computational pipelines have been integrated to support end-to-end analysis across organizational private datasets, shared consortia data, and public repositories, including NIST--JARVIS, the Materials Project, and related national data infrastructures. These pipelines operate natively within the platform’s data ingestion layer and AI-driven application stack, enabling researchers to combine experimental data with machine-learning surrogates and computational results within a unified environment. By reducing the overhead associated with data access, transformation, and workflow construction, this integration allows scientific teams to concentrate effort on downstream materials design and decision-making.

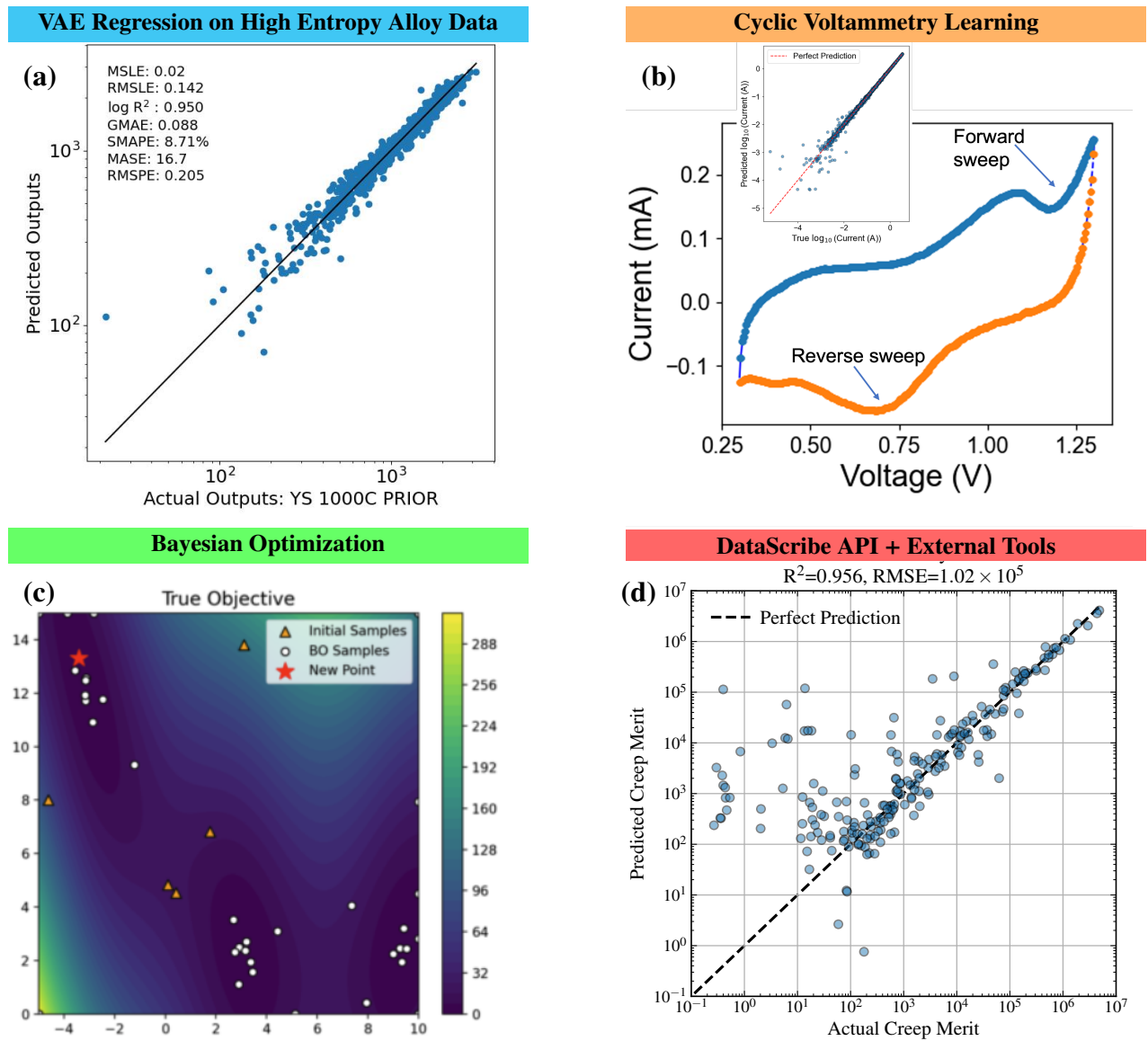
\begin{figure}[!ht]
    \centering
    % ---------- Panel (a) ----------
    \begin{overpic}[width=0.45\linewidth]{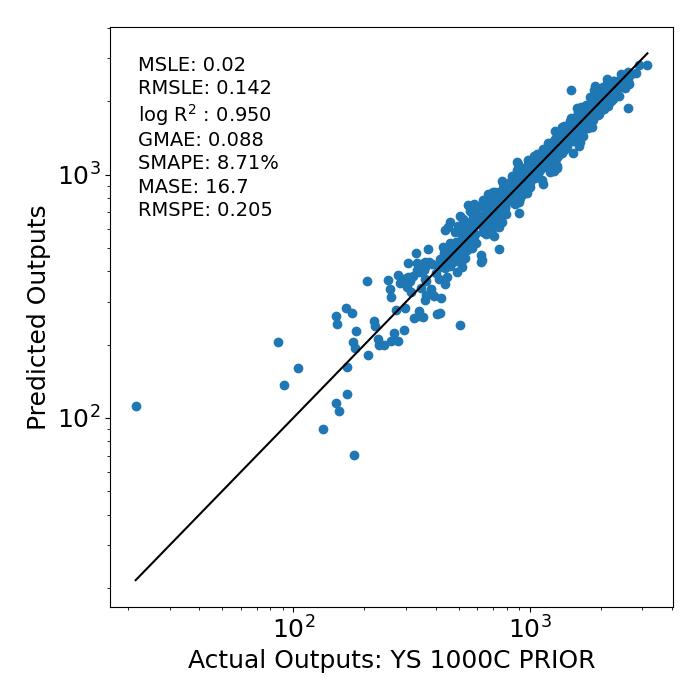}
        % Colored bar
        \put(0,99){\colorbox{cyan!60}{\parbox{0.45\columnwidth}{\centering \textbf{VAE Regression on High Entropy Alloy Data}}}}
        % Panel label
        \put(3,88){\large\textbf{(a)}}
    \end{overpic}
    \hfill
    % ---------- Panel (b) ----------
    \begin{overpic}[width=0.45\linewidth]{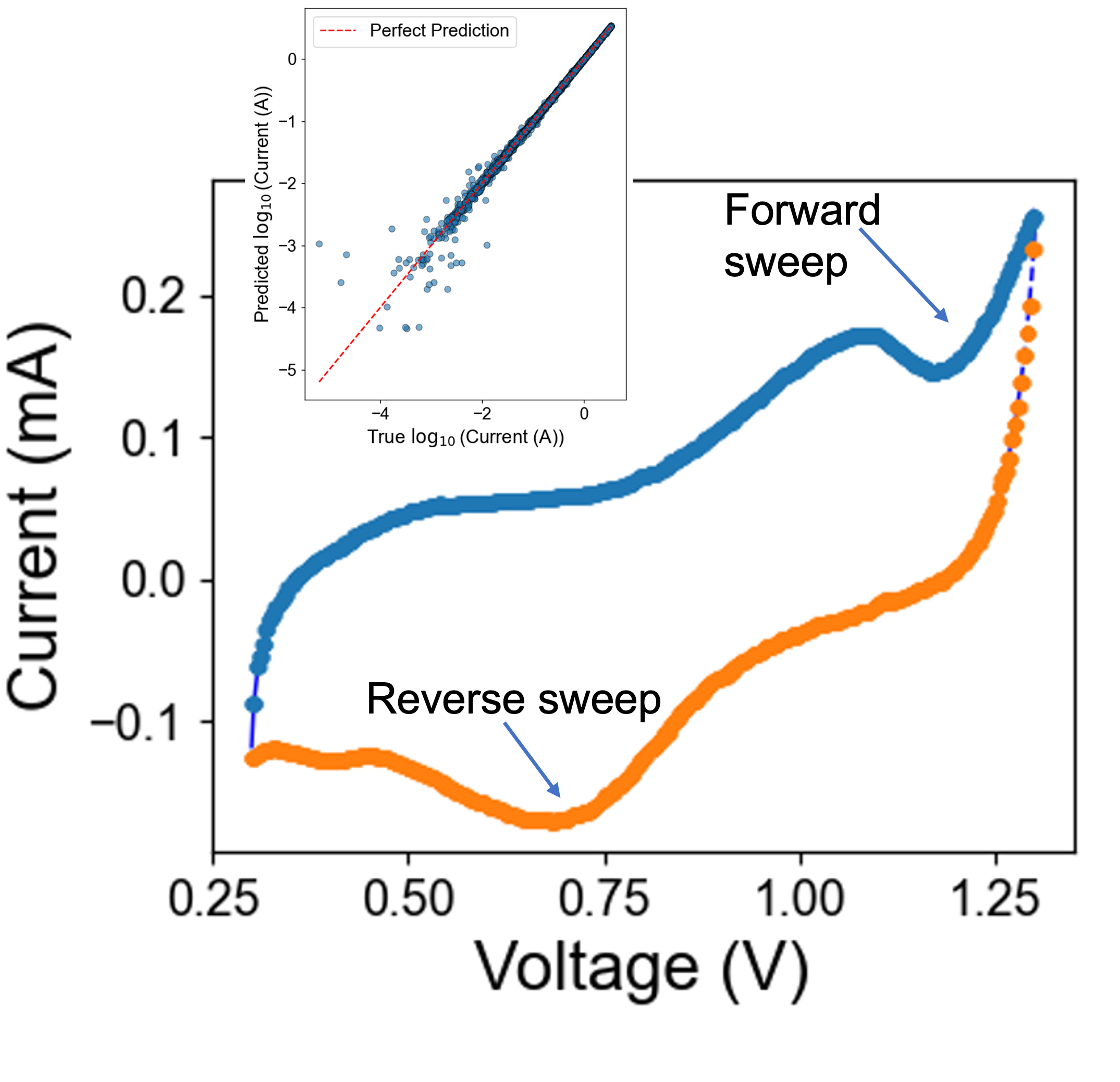}
        \put(0,99){\colorbox{orange!60}{\parbox{0.45\columnwidth}{\centering \textbf{Cyclic Voltammetry Learning}}}}
        \put(3,88){\large\textbf{(b)}}
    \end{overpic}
    
    \vspace{0.5em}
    
    % ---------- Panel (c) ----------
    \begin{overpic}[width=0.45\linewidth]{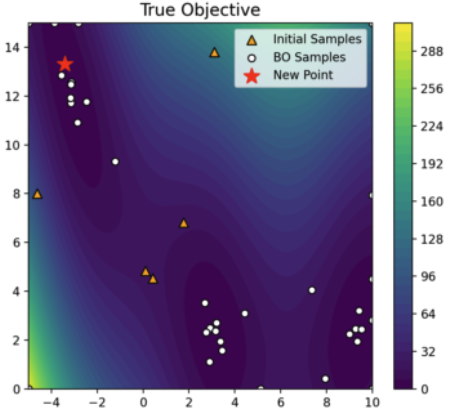}
        \put(0,98){\colorbox{green!60}{\parbox{0.45\columnwidth}{\centering \textbf{Bayesian Optimization}}}}
        \put(3,88){\large\textbf{(c)}}
    \end{overpic}
    \hfill
    % ---------- Panel (d) ----------
    \begin{overpic}[width=0.45\linewidth]{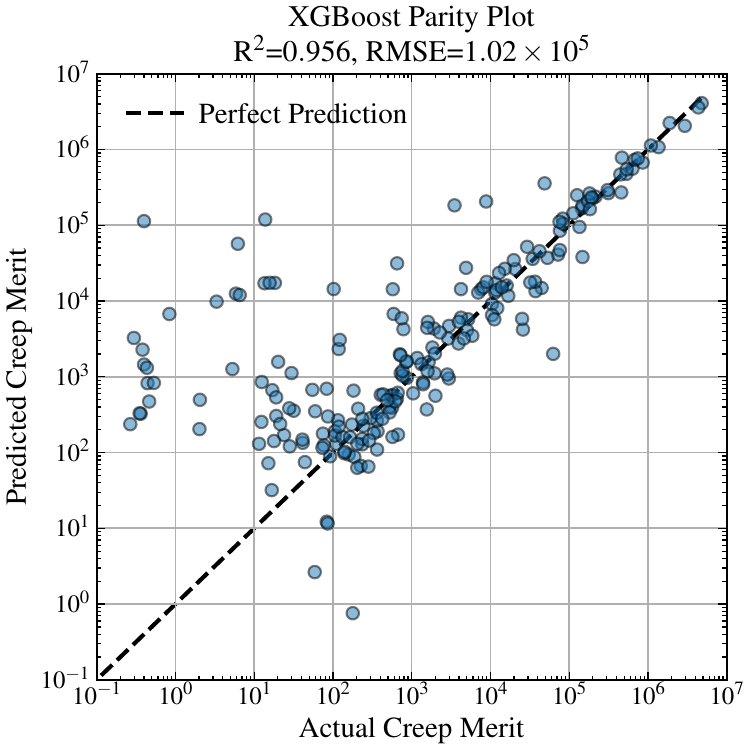}
        \put(0,98){\colorbox{red!60}{\parbox{0.45\columnwidth}{\centering \textbf{DataScribe API + External Tools}}}}
        \put(-1,88){\large\textbf{(d)}}
    \end{overpic}

    \caption{(a) Result of Variational Encoder-Decoder Regressor trained on Refractory High Entropy Alloy Dataset showing Log-Log parity plot of predicted vs actual yield strength at 1000$^\circ$C, (b) Result obtained from training Vanilla Encoder-Decoder Regressor on Cylic Voltammetry Data showing the predicted CV forward and inverse loops (markers) overlaying the experimental values (solid line) The plot on top corner highlights the parity plot. (c) Result showing the obtained Bayesian Query after 35 iterations based on 5 initial random samples over Goldstein-Price function, a smooth multimodal function with three global minima, (d) Result obtained by invoking DataScribe API and integration of XGBoost and ScikitLearn showing Log–log parity plot of predicted vs\. actual Creep Merit for a held‑out test set from an XGBoost regressor trained on Nb, Cr, V, W, and Zr}
    \label{fig:example-Scientific-Results}
\end{figure}

DataScribe currently supports three in-house, production-ready workflows. The first comprises a family of deep-learning regression models based on encoder--decoder architectures for tabular data, including standard encoder--decoder networks, variational encoder--decoder models, and DNNF-based variants. Users can configure architectural parameters and evaluate these models using either proprietary datasets or curated datasets available within the platform. A second workflow focuses on redox-state prediction from cyclic voltammetry, magnetic hysteresis, and other loop-type measurements, employing Random Forest and CatBoost models. In addition to these pipelines, a Bayesian multi-objective, multi-fidelity optimization workflow based on deep Gaussian process regression is natively deployed and accessible to platform users. Educational workflows have also been implemented for demonstration and benchmarking, including a single-objective Bayesian optimization pipeline for minimizing standard analytical test functions such as Branin, Goldstein--Price, and Schwefel. These workflows integrate data preprocessing, model selection, and uncertainty quantification, serving both as turnkey analytical tools and as reference implementations for the development and extension of custom workflows. Representative calculations for each workflow are shown in Fig.~\ref{fig:example-Scientific-Results}.

%%%%
%%%%
\subsection{Case Study 1: Encoder-Decoder Regression Workflow on High Entropy Alloy Data}
%3.1 Case Study 1 — Deep-Learning Regression on Materials Data
%Move the encoder–decoder workflow here.

The \emph{Encoder--Decoder} workflow (Fig.~\ref{Sfig:encoder-decoder-workflow}) implements a suite of deep neural architectures for nonlinear regression on tabular materials datasets, using an encoder to map input features into either high-dimensional (overcomplete) or lower-dimensional (undercomplete) latent representations and a decoder to reconstruct target properties. The workflow provides comprehensive preprocessing options, including quantile transformation, log1p, sigmoid, and min--max scaling, with side-by-side kernel density plots to visualize the effects of each transformation, as well as optional threshold-based filtering. Architecture parameters are fully configurable, including layer counts, neuron distributions, and latent dimensionality. During training, step-decay learning rate scheduling is employed, and the system automatically generates diagnostic outputs, including training loss histories and parity plots in both scaled and original feature spaces, with logarithmic scaling to assess convergence and predictive performance across the full dynamic range of target properties. Trained models are immediately deployable through an interactive prediction interface in which users specify custom feature values, with the workflow applying the full preprocessing pipeline and inverse-transforming predictions to physical units. This enables rapid computational screening of hypothetical material compositions, with all results available for export. The workflow structure and key elements of the graphical user interface are shown in Fig.~\ref{Sfig:encoder-decoder-workflow}.

%%%%
%%%%
%%%%
%%%%
\subsection{Case Study 2: Redox Prediction from Cyclic Voltammetry Data}
%3.2 Case Study 2 — Redox Prediction from CV Data
%Move the RedoxPredict workflow here.

The \emph{Redox Predict} workflow (Fig.~\ref{Sfig:redox-predict-workflow}) applies machine-learning regression models, including CatBoost and Random Forest, to predict current--voltage responses in cyclic voltammetry experiments across varying scan rates, materials, and sweep directions, addressing limitations of traditional peak-fitting approaches in complex multi-electron or irreversible systems. The workflow constructs current--voltage datasets from wide-format cyclic voltammetry tables by mapping material-specific properties (e.g., hydrogen desorption barriers and surface coverage) through an interactive editor, automatically identifying voltage--current pairs, segmenting forward and reverse sweeps, and assembling regression-ready feature matrices with optional scaling transformations. Model training employs three-way data splitting with early stopping, generating diagnostic outputs that include RMSE convergence curves, logarithmic parity plots with $R^2$ metrics, and predicted-versus-experimental cyclic voltammetry overlays that demonstrate reconstruction of characteristic redox features. Post-training analysis supports matrix-based visualizations spanning multiple materials and scan rates for systematic validation, a hypothetical prediction interface for computational screening of untested compositions, and SHAP-based feature importance analysis that quantifies the contributions of voltage, scan rate, and material properties to model predictions. All outputs are exportable for downstream analysis, enabling interpretation of structure--property relationships and supporting mechanistic insight.

\subsection{Case Study 3: Multi-Objective Bayesian Optimization}
%3.3 Case Study 3 — Multi-Objective Bayesian Optimization
%Move MOMF and BO results here.

\begin{figure}[ht!]
    \centering
    \includegraphics[width=0.95\linewidth]{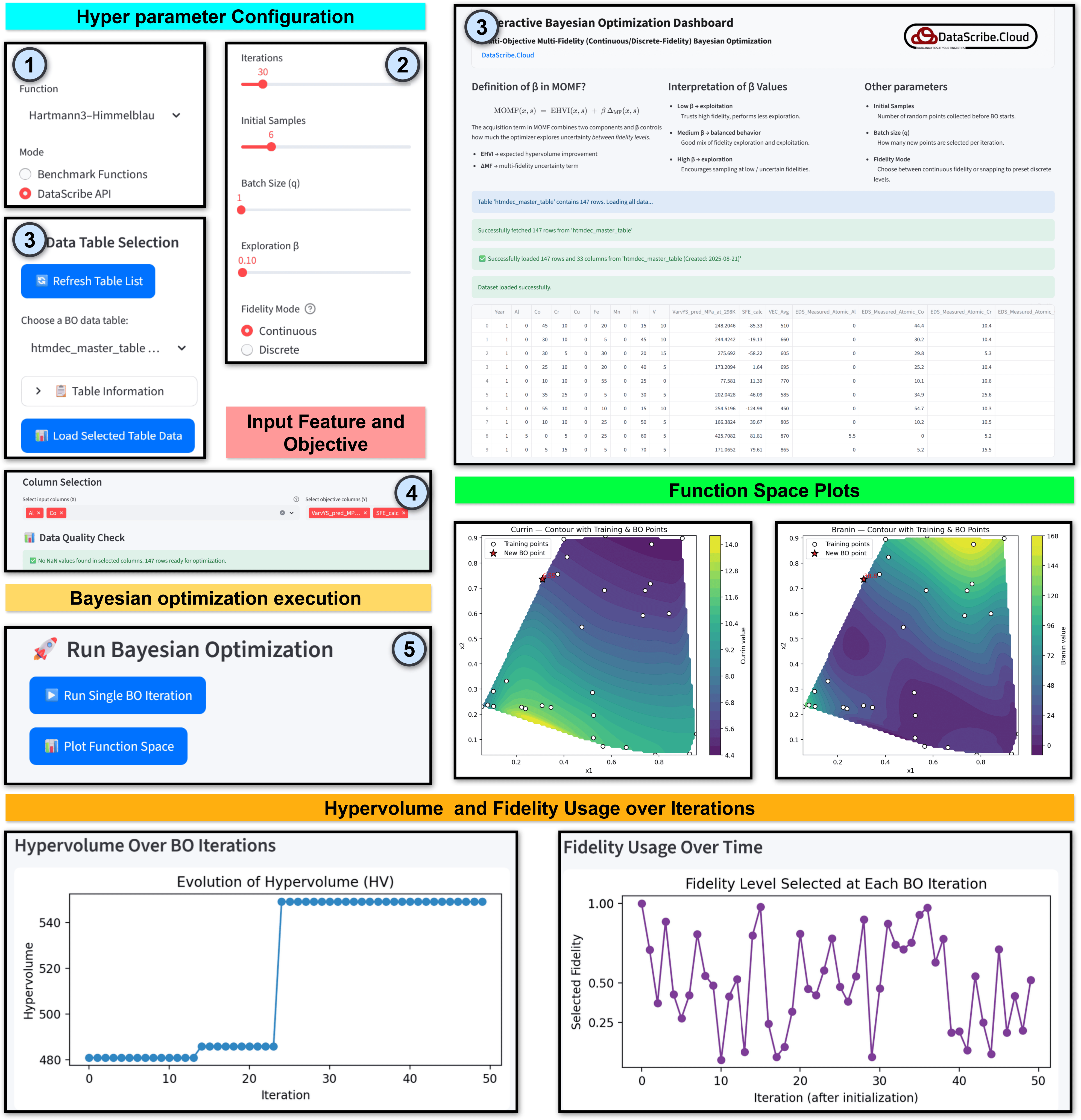}
    \caption{DataScribe Bayesian Multi-Objective Multi-Fidelity Optimization workflow. \textbf{Step 1:} Function and mode selection (benchmark functions or DataScribe API). \textbf{Step 2:} Hyperparameter configuration including iteration count, initial samples, batch size ($q$), exploration parameter ($\beta$), and fidelity mode. \textbf{Step 3:} Data table selection from cloud platform with automatic metadata retrieval and NaN handling. \textbf{Step 4:} Input feature ($X$) and objective ($Y$) column specification with fidelity configuration. \textbf{Step 5:} Sequential Bayesian optimization execution generating newly selected points with acquisition values, hypervolume metrics, and function space visualization showing observed points, Pareto front, and optimization trajectory. Additional diagnostic outputs include fidelity usage distribution, fidelity selection over iterations, hypervolume evolution, hypervolume improvement ($\Delta$HV), distance-to-Pareto-front convergence, and acquisition value progression across iterations. All results exportable as CSV files for downstream analysis.}
    \label{fig:bayesian-mobo-workflow}
\end{figure}

The \emph{Bayesian Multi-objective Multi-fidelity (MOMF) Optimization} workflow (Fig.~\ref{fig:bayesian-mobo-workflow}) implements an active learning framework for navigating high-dimensional materials design spaces under competing performance objectives with heterogeneous data fidelity arising from mixed experimental and computational evaluations. In this setting, surrogate models are iteratively trained on accumulated data, and new experimental or computational queries are selected sequentially or in batch using acquisition functions that balance performance improvement, uncertainty reduction, and resource expenditure. This design enables systematic navigation of high-dimensional design spaces while minimizing the number of expensive evaluations required to identify high-performing trade-off solutions. 

The workflow supports two operational modes: benchmark testing using standard multi-objective test problems (e.g., Branin--Currin, Booth--Rastrigin, Hartmann--Himmelblau) for algorithm validation, and integration through the DataScribe API for optimization on user-provided experimental datasets with automated data quality assessment, including NaN detection and multiple imputation strategies. Users configure key hyperparameters such as the exploration--exploitation trade-off parameter $\beta$, batch size $q$ for parallel candidate selection, and fidelity mode (continuous or discrete) to accommodate varying computational or experimental cost structures. 

At the core of the workflow is a multi-objective Bayesian optimization loop in which Gaussian process or surrogate models approximate multiple objective functions simultaneously, capturing predictive uncertainty across the design space. Candidate points are selected using an acquisition strategy based on expected hypervolume improvement (EHVI) in the sequential setting, or q-expected hypervolume improvement (qEHVI) in the batch setting, such that candidate designs are obtained by maximizing the acquisition function,
\begin{equation}
\mathbf{X}_{\mathrm{new}} = \arg\max_{\mathbf{X} \subset \mathcal{X}} \, \alpha_q(\mathbf{X}),
\end{equation}
where $\alpha_q(\cdot)$ denotes the batch acquisition function (with $q=1$ for the sequential EHVI case), which quantifies the expected gain in Pareto-optimal objective space resulting from one or more new evaluations. By explicitly optimizing hypervolume, the workflow naturally accounts for trade-offs among competing objectives rather than collapsing them into a single scalar metric.

A distinguishing feature of the workflow is its support for multi-fidelity optimization. Different fidelity levels corresponding, for example, to low-cost surrogate evaluations, medium-fidelity simulations, or high-fidelity experimental measurements are treated as alternative information sources with distinct uncertainty and cost characteristics. Fidelity can be represented either as a continuous variable or as a discrete categorical input, allowing flexible modeling of hierarchical evaluation strategies commonly encountered in materials science. 

Cost awareness is incorporated at the benchmarking stage by embedding a cost parameter directly in the acquisition function, and in application mode through fidelity-dependent cost ratios. Each fidelity level can be assigned a user-defined cost ratio reflecting relative computational or experimental expense, and optimization campaigns can be executed under fixed evaluation budgets rather than fixed iteration counts. These cost assignments influence acquisition decisions by encouraging exploration at lower-fidelity, lower-cost levels during early stages of high epistemic uncertainty, followed by selective refinement using higher-fidelity evaluations as the Pareto front stabilizes. This approach enables realistic assessment of optimization performance under resource constraints and aligns the workflow with practical decision-making scenarios in laboratory and high-performance computing environments.

Each optimization iteration returns a ranked set of recommended design points together with predicted objective values, acquisition scores, fidelity selections, and current hypervolume metrics. To support interpretability and diagnostic assessment, the workflow generates a suite of visual outputs, including Pareto front evolution, hypervolume and incremental hypervolume improvement ($\Delta$HV) curves, fidelity usage distributions, and distance-to-Pareto-front convergence traces. Interactive function-space visualizations further illustrate the spatial relationship between observed samples, candidate queries, and current Pareto-optimal regions. All intermediate and final results are exportable as structured CSV files, facilitating downstream analysis and integration with experimental planning workflows.

%\subsection{Single-Objective Bayesian Optimization}

The sequential Bayesian Optimization workflow (Fig.~\ref{Sfig:single-obj-bo-workflow}) provides an interactive educational environment for exploring Gaussian process--based optimization on standard benchmark functions. Users select from a set of classical test problems spanning diverse landscape characteristics, including smooth multimodal surfaces (Branin), highly oscillatory functions (Ackley, Rastrigin), functions with distant global minima (Schwefel), and rugged landscapes with sharp gradients (Eggholder), each accompanied by descriptive annotations outlining the associated optimization challenges. Three acquisition functions are supported: expected improvement (EI) for balanced exploration and exploitation, probability of improvement (PI) for exploitation-focused search, and lower confidence bound (LCB) with an adjustable $\beta$ parameter for user-controlled exploration weighting. During optimization, four-panel visualizations update iteratively to display the true objective surface, the Gaussian process posterior mean, predictive uncertainty, and the acquisition function landscape with sample trajectories overlaid, enabling direct inspection of surrogate model evolution and sequential sampling behavior. Post-optimization diagnostics report the best observed objective value together with distance to the known global optimum, while additional panels present convergence curves, step-size evolution characterizing exploration behavior, and acquisition values across iterations. All optimization histories are exportable as CSV files for offline analysis and pedagogical use.

% \textcolor{blue}{[MOve the text and Explain the result obtained by using datascribe API]}

% \textcolor{red}{\Cref{fig:example-Scientific-Results}(d) is a log–log parity plot showing predicted versus actual Creep Merit for a held‑out test set from an XGBoost regressor trained on Nb, Cr, V, W and Zr; each data point is a sample and the dashed diagonal marks perfect prediction. The tight clustering of points around the diagonal across several orders of magnitude, together with R$^{2}$ = 0.956 and RMSE $\approx$ \num{1.02e5}, indicates strong agreement and that the model captures most of the variance, though slightly increased scatter at the distribution tails suggests under/over‑prediction for extreme values. This evaluation and visualization were made possible by retrieving the table directly with the \emph{datascribe\_api} Python client.}

\subsection{Integration with External Computational Tools}

Beyond DataScribe’s native workflows, the platform’s \emph{datascribe\_api} Python client~\cite{datascribe_api_2025} enables integration with external machine-learning frameworks, allowing researchers to programmatically retrieve structured datasets and apply established tools such as XGBoost, scikit-learn, and PyTorch. This interface supports the development of custom analytical pipelines while retaining access to DataScribe’s unified data management and governance layer. For example, datasets retrieved through the API can be passed directly to gradient-boosting regressors for property prediction tasks in alloy design studies.

~\ref{Sapp:datascribe_xgboost} in the Supplementary Document presents a complete, reproducible example demonstrating XGBoost regression applied to refractory high-entropy alloy datasets, achieving strong predictive performance ($R^2 = 0.956$) across multiple orders of magnitude in creep merit values. This example illustrates how DataScribe operates not as a closed platform, but as an extensible data backbone that integrates cleanly with the broader Python scientific computing ecosystem. In this setting, researchers can apply domain-specific modeling approaches while retaining the benefits of centralized data stewardship, structured metadata management, and reproducible analysis workflows.

These pre-built workflows and external integration capabilities reflect a design approach that emphasizes accessibility alongside methodological rigor, allowing domain experts to apply contemporary machine-learning and optimization techniques without requiring specialized expertise in algorithm development or software engineering. As the platform matures, the visual workflow canvas is intended to evolve from a planning and visualization interface into a fully executable environment, enabling researchers to construct arbitrarily complex, reproducible analytical pipelines aligned with specific materials discovery objectives.

%% file: 02_1_datascribe-architecture-and-integrations.tex
\section{Architecture and Intelligence Stack} %% Divyanshu - Vahid

\begin{figure}[!ht]
\centering
\begin{overpic}[width=0.95\linewidth,trim=0 0 0 40,clip]{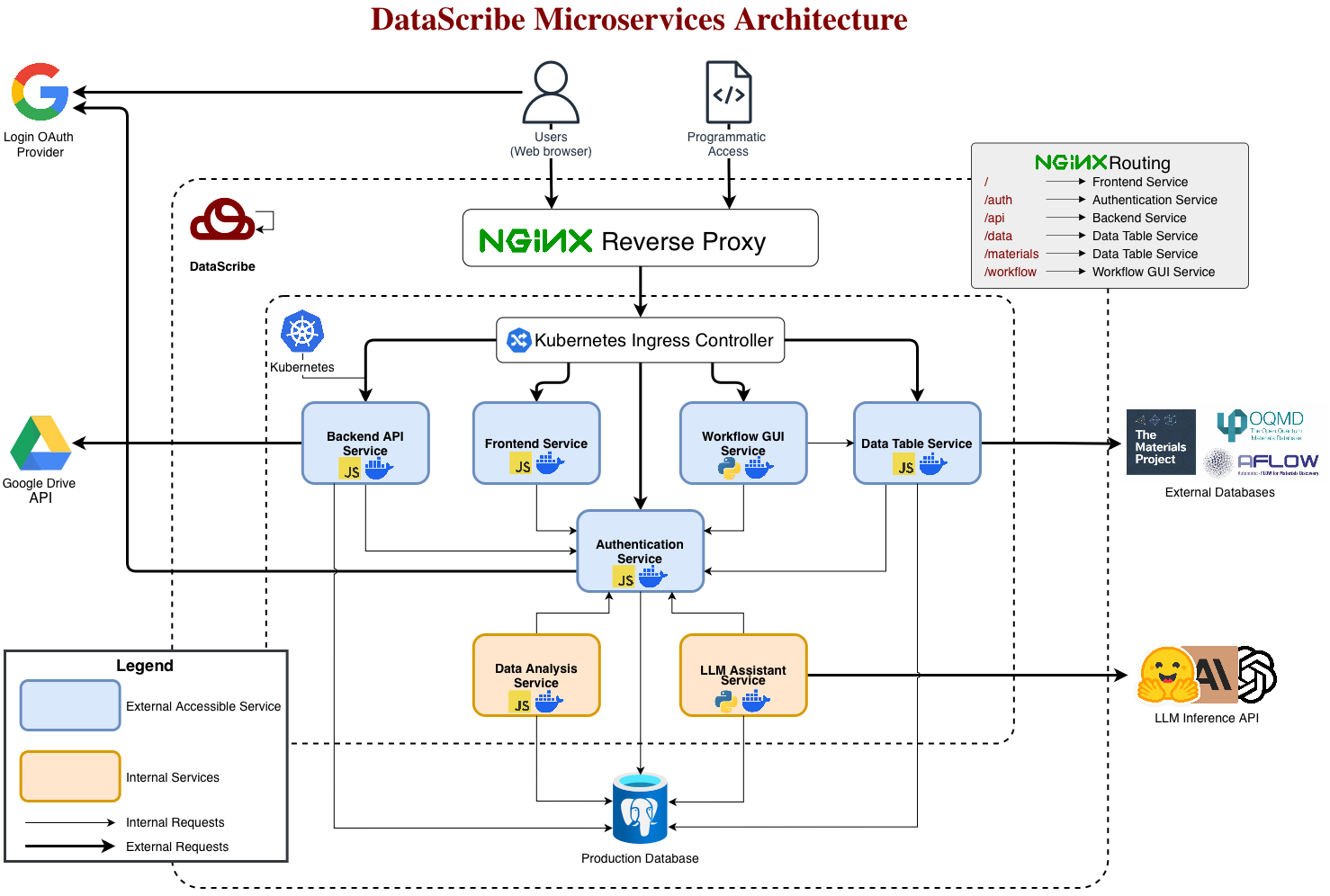}
    % --- Title box overlay ---
    \put(25,65){\colorbox{green!60}{\parbox{0.50\columnwidth}{\centering \textbf{DataScribe Microservices Architecture}}}}
\end{overpic}
\caption{DataScribe Microservices Architecture. The system employs a distributed architecture with NGINX reverse proxy handling external requests, Kubernetes Ingress Controller managing internal routing, and seven core microservices: Backend API Service, Frontend Service, Data Table Service, Workflow GUI Service, Authentication Service, Data Analysis Service, and LLM Assistant Service. External integrations include OAuth providers (currently Google OAuth), Google Drive API, LLM inference APIs (Hugging Face), external materials databases (OQMD, Materials Project, AFLOW), and a production database for persistent storage. All services are containerized and orchestrated within a Kubernetes cluster for scalability and resilience.}
\label{fig:schematic-architecture}
\end{figure}

DataScribe is implemented as an AI-native research platform whose architecture is designed to support closed-loop materials discovery. The system is structured as a collection of modular microservices that coordinate data ingestion, workflow execution, analysis, and intelligent assistance. Rather than prioritizing software complexity, the architectural choices emphasize acceleration of the scientific cycle by linking data capture, interpretation, and decision-making in a reproducible and scalable manner. Figure~\ref{fig:schematic-architecture} summarizes the platform architecture and illustrates how DataScribe’s services maintain continuity across experimental workflows, simulation pipelines, and AI-driven inference.

The platform supports cloud-based operation by default, with persistent storage currently integrated through Google Drive and additional providers accessible via pluggable connectors. Although not presently coupled to a high-performance computing (HPC) cluster, the same API- and metadata-driven architecture supports straightforward extension to HPC centers and automated laboratory environments as needed. This design enables the platform to accommodate increasing campaign complexity, ranging from exploratory model development to high-throughput, closed-loop optimization, without disrupting established scientific workflows. In this configuration, the architecture provides an extensible foundation for multimodal data integration, coordinated agent-based execution, and sustained, traceable experimentation.

%DataScribe adopts a modular microservices architecture that cleanly separates backend logic, frontend clients, data services, analytical services, workflow services, and AI-driven assistant components. As illustrated in Fig.~\ref{fig:schematic-architecture}, these microservices communicate through stable API contracts while remaining independently deployable and scalable. This design emphasizes robustness, extensibility, and long-term maintainability, allowing components such as ingestion pipelines, metadata services, workflow engines, and analytical modules to evolve without disrupting the broader ecosystem. Integration across heterogeneous data sources, including computational simulations, and literature-derived datasets is enabled through RESTful APIs, ontology-backed metadata standards, and role-based access control. This ensures interoperability with external cyberinfrastructure while maintaining robust security and provenance tracking. 

%DataScribe is purpose-built for collaborative materials acceleration. Its API-based architecture enables coupling between autonomous experiments, high-performance computing pipelines, and multi-institutional data ecosystems, providing a flexible foundation for reproducible, large-scale discovery campaigns. This distributed design not only addresses current integration challenges but also anticipates future requirements, including interoperability with self-driving laboratories and AI-driven agents, positioning DataScribe as an extensible framework for next-generation materials design and discovery.

%%%%
\subsection{Microservices}

DataScribe employs seven modular microservices that support closed-loop materials discovery. These services are illustrated in Fig.~\ref{fig:schematic-architecture}. By separating data ingestion, analysis, workflow orchestration, and intelligent assistance into independently deployable components, the platform is designed to maintain a stable and extensible interface across the discovery cycle. Each microservice corresponds to a functional stage within the propose--measure--learn loop, ensuring that data, models, and decision logic propagate consistently across experimental workflows, simulation pipelines, and AI-driven agents.

The platform organizes these microservices into two logical layers. Externally accessible services, exposed through an NGINX reverse proxy, support secure browser-based interaction as well as programmatic access by automated tools and agents. Internal services execute within the Kubernetes cluster and are not directly exposed to end users, handling sensitive computation, data transformation, and large language model--assisted reasoning. Kubernetes provides automated scaling, failure isolation, and reliable service-to-service routing, ensuring that high-throughput workloads do not degrade overall system performance.

The \textbf{(1)~Backend API Service} serves as the orchestration layer, exposing stable endpoints through which experiments, simulations, workflows, and agents interact. The \textbf{(2)~Frontend Service} provides a responsive, browser-based environment for exploration of datasets, workflows, and analytical results. The \textbf{(3)~Workflow GUI Services} enable node-based construction and execution of analysis pipelines, allowing users to launch MOBO campaigns, data transformations, or surrogate-model evaluations without writing code. The \textbf{(4)~Data Table Service} manages heavy tabular operations, including large structure–property matrices and high-throughput computational outputs, without compromising system responsiveness.

Two internal compute services support scalable analysis: the \textbf{(5)~Data Analysis Service}, which executes ML pipelines, scientific Python routines, surrogate models, and batch evaluations; and the \textbf{(6)~LLM Assistant Service}, which integrates external LLM inference APIs to support natural-language querying, metadata enrichment, literature-aware reasoning, and AI agent interaction. The \textbf{(7)~Authentication Service} ensures secure access across institutions using federated identities (currently Google OAuth), enabling governed collaboration while maintaining data protection. These microservices form a cohesive and extensible architecture. Each component can scale independently, update without disturbing downstream workflows, and integrate with cloud, laboratory, or future HPC resources. This layered design, external access, internal computation, and Kubernetes orchestration provides the reliability, security, and flexibility required for sustained, autonomous materials discovery campaigns. Additional implementation details including runtime frameworks, deployment tools, and cloud components—are provided in Supplementary Table ~\ref{Stab:tech_stack}.

\subsection{Integration across lab, cloud, and HPC}
DataScribe currently operates as a cloud-first platform, using cloud-based compute and storage resources as a unified execution fabric for analytics, workflow orchestration, and AI-assisted reasoning. Organizations retain control over their data by connecting DataScribe to a selected cloud storage provider and exposing only the datasets required for a given workflow. The architecture is intentionally extensible, with shared API and metadata layers that support future integration of laboratory instrumentation, autonomous experiment controllers, and institutional or national high-performance computing (HPC) systems without requiring workflow reconfiguration. This design allows propose--measure--learn cycles to scale from small exploratory analyses to high-throughput computational screening and automated experimental campaigns. Even in the absence of direct laboratory or HPC connectivity, the platform is structured to support closed-loop discovery across heterogeneous computational and experimental environments.

\subsection{Security, provenance, and governance}

Security and provenance are integral to the design of DataScribe, ensuring that autonomous and multi-agent workflows remain trustworthy and reproducible. Role-based access control and federated authentication support secure collaboration across institutional boundaries while preserving fine-grained control over data access. Each dataset, workflow execution, and model update is assigned a persistent UUID accompanied by comprehensive lineage metadata, including data sources, transformations, computational context, and environmental parameters, enabling traceability from ingestion through deployment. FAIR-compliant metadata schemas, audit trails, and versioned datasets and pipelines further reinforce reproducibility and accountability. These governance mechanisms provide a stable and regulated substrate for sustained, AI-driven discovery workflows spanning academic, industrial, and multi-institutional materials acceleration platforms.

%% file: 02_2_heterogeneous-data-unification-workflow.tex
\section{Heterogeneous Data Unification}

\begin{figure}[!ht]
    \centering
    \includegraphics[width=0.85\linewidth]{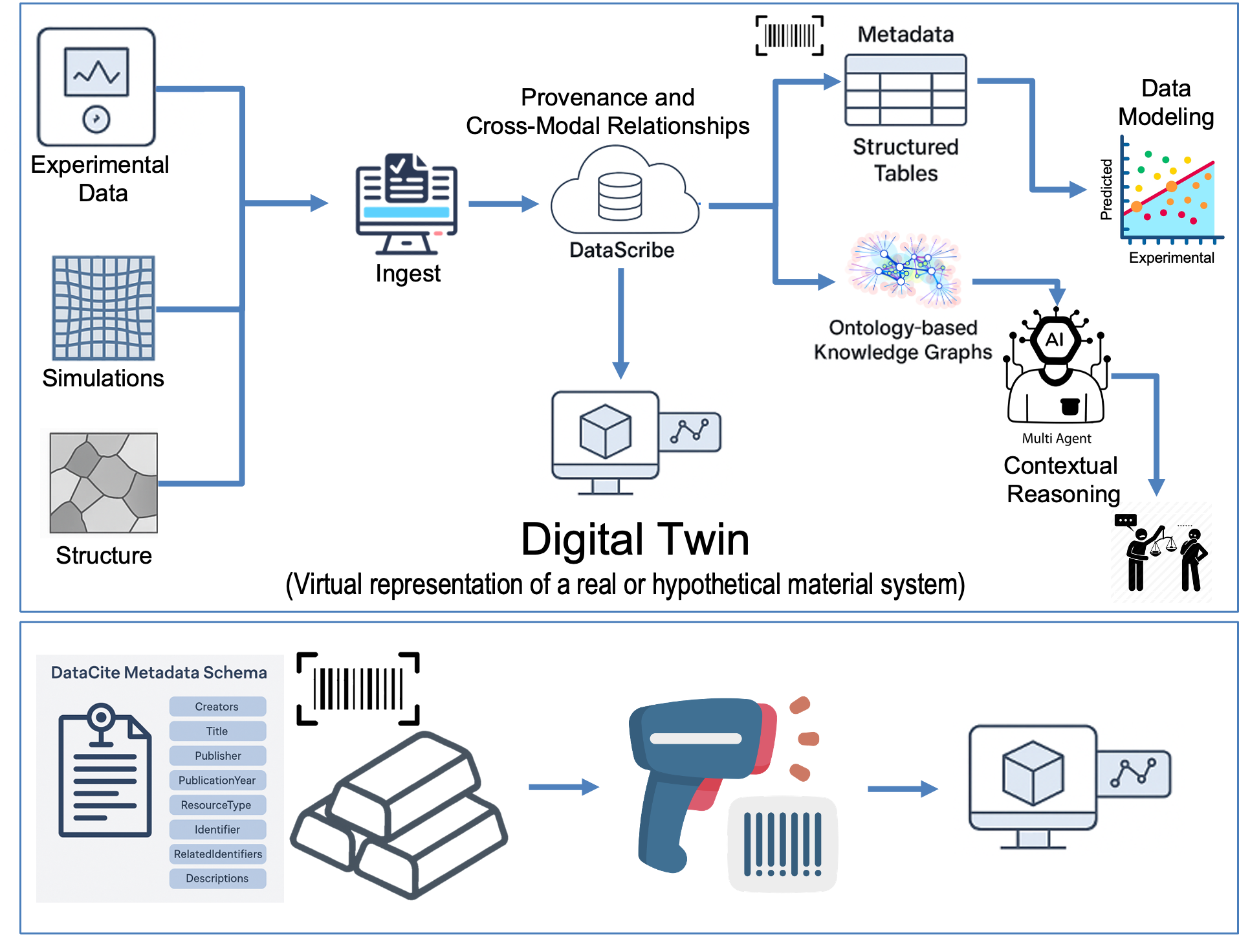}
    \caption{\textbf{DataScribe-enabled digital-twin generation from heterogeneous materials data.} 
    Experimental data, simulations, and structural information are ingested and semantically enriched through provenance tracking and cross-modal relationships within DataScribe. These unified datasets populate structured tables and ontology-based knowledge graphs, enabling metadata integration, contextual multi-agent reasoning, and downstream data modeling. The aggregated knowledge representation forms the basis of a digital twin—a virtual counterpart of a real or hypothetical materials system. The lower panel illustrates how physical samples enter this workflow through barcode identification and automated ingestion, linking laboratory processes directly to AI-driven modeling and digital-twin construction.}
    \label{fig:datascribe-digital-twin}
\end{figure}

Many of the capabilities currently emphasized, such as unit normalization, schema mapping, FAIR metadata capture, and structured workflow or “traveler” forms, are standard features found across mature laboratory informatics systems and commercial materials data platforms. Established Eelctronic Lab Notebooks (ELNs) often are researcher-facing and document the experiments whereas Laboratory Information Management Systems (LIMS) are operations-facing and track samples, workflows, QA/QC (operations-facing). ELNs such as \textbf{Benchling}, \textbf{LabArchives}, and \textbf{ChemOS}~\cite{sim2024chemos} for chemical self-driving laboratories already provide schema-on-write data models, controlled vocabularies, provenance tracking, and form-based data entry. LIMS such as \textbf{LabWare}, \textbf{STARLIMS}, and \textbf{Thermo Fisher SampleManager} similarly offer structured metadata management, instrument integration, audit trails, and workflow orchestration for regulated or industrial environments. Materials-focused platforms including Citrination, Materials Commons, AiiDA, and Materials Project’s MAPI expose similar mechanisms for metadata standardization, identifier management, and cross-format ingestion. Even industrial data fabrics such as Palantir Foundry or Databricks support ontology mapping, unit normalization, and FAIR-aligned metadata layers at scale. Because these capabilities are now broadly commoditized across scientific informatics ecosystems, highlighting them as core innovations risks obscuring the genuine novelty of DataScribe and diminishing its differentiation in a high-profile research context. In contrast to these schema-on-write–centric systems, DataScribe provides an AI-native data substrate (Fig.~\ref{fig:datascribe-digital-twin}) that integrates a hybrid relational--graph semantic layer engineered specifically for multi-objective optimization, agentic workflows, and model--experiment iteration. Incoming data streams from experimental, computational, and literature sources are contextualized through dynamically extensible ontologies that capture provenance, hierarchical relationships, and cross-modal dependencies among CALPHAD grids, CV curves, DFT calculations, microstructure images, and process logs. This semantic substrate enables autonomous agents to perform contextual ingestion, metadata enrichment, schema inference, and linkage of heterogeneous datasets into a unified knowledge representation that feeds directly into Bayesian optimization, uncertainty-aware modeling, and digital-twin construction. While the current platform provides strong governance and reproducibility, its schema-on-write foundations still limit flexibility for emerging multimodal and adaptive workflows; future extensions will adopt schema-on-read strategies and richer graph-based representations to capture evolving metadata, agentic provenance, and adaptive experiment logic with greater expressiveness and fidelity.

%% file: 02_4_datascribe-data-management-interface.tex
\section{DataScribe Organization and Data Management Layers}

DataScribe implements a role-based organizational model that supports secure collaboration across institutions, research groups, and projects. Administrators manage organizational hierarchies and permissions, while project-level roles govern data access, provenance capture, and workflow deployment. Beyond access control, the platform provides a unified environment for ingesting, organizing, and semantically enriching heterogeneous datasets through structured metadata, schema validation, and ontology-based mapping (Fig.~\ref{Sfig:data-management-framework}). These management layers ensure that experimental, computational, and literature-derived data are consistently represented, traceable across user and workflow transitions, and immediately interoperable with the semantic and analytics layers that drive autonomous modeling and digital-twin construction. Extended descriptions of the data repository, file formats, metadata structures, and organizational workflows are provided in the Supplementary Information.

%% file: 07_apis.tex
\section{APIs and Programmatic Interfaces} \label{sec:api-programmatic-interfaces}

\begin{figure}[h]
    \centering
    \includegraphics[width=0.75\linewidth]{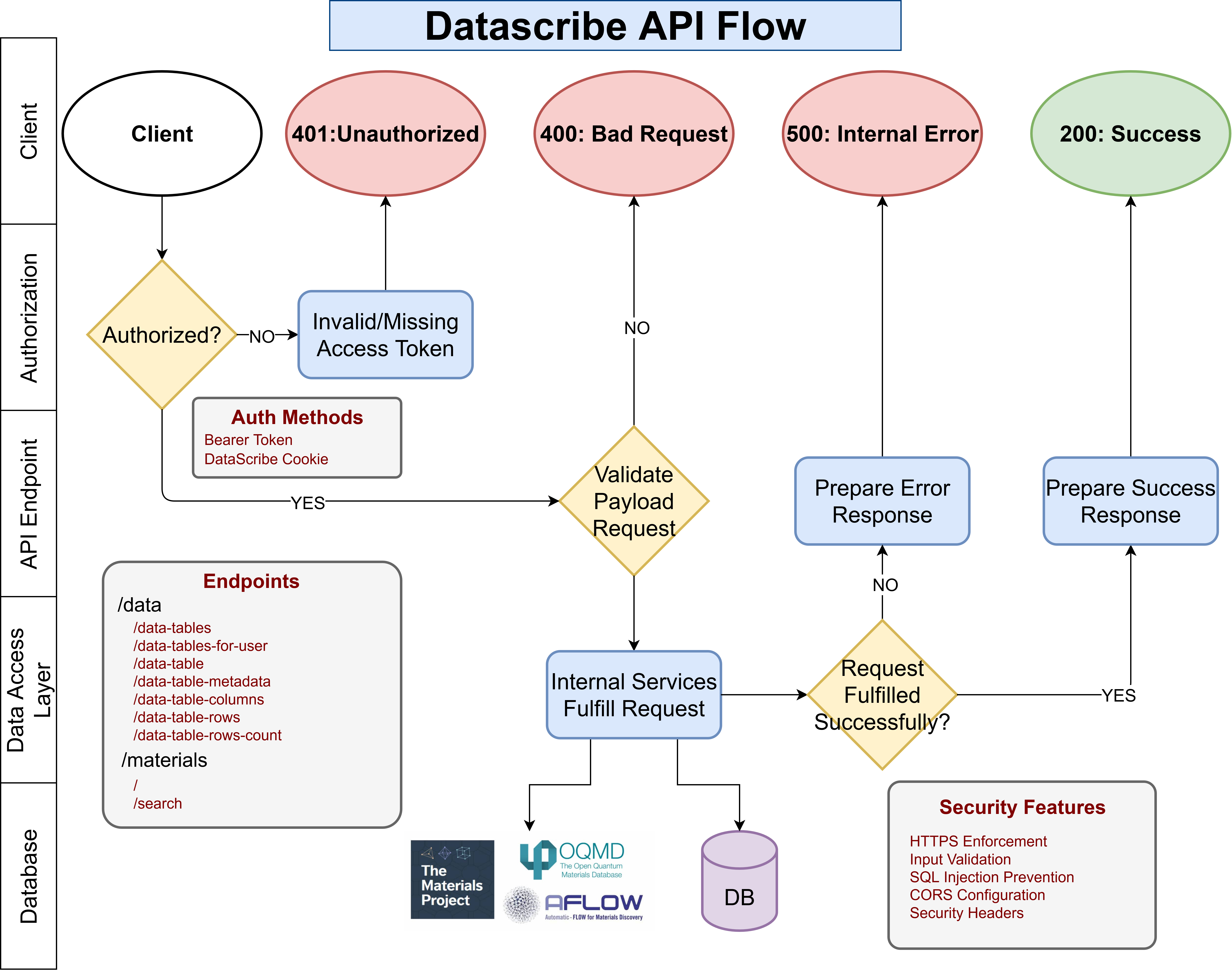}
    \caption{DataScribe API Request Processing Flow. The architecture shows the four-layer request lifecycle: client requests pass through OAuth authentication, payload validation, internal service processing, and database interaction. Error responses (401, 400, 500) and successful responses (200) are returned based on the processing outcome at each layer.}
    \label{fig:datascribe-api-flow}
\end{figure}

DataScribe provides a unified programmatic interface for automated data ingestion, querying, and materials database search, enabling integration with external workflows and computational pipelines. A RESTful API layer (Fig.~\ref{fig:datascribe-api-flow}) exposes structured access to datasets, schema metadata, and row-level queries, while a unified materials search interface aggregates results from the Materials Project, AFLOW, and OQMD into a consistent representation suitable for cross-database analysis. Programmatic permissions and access controls are aligned with organizational roles, allowing secure retrieval of experimental records, computed properties, and curated materials datasets within automated pipelines supporting Bayesian optimization, surrogate modeling, and digital-twin construction. In this configuration, data retrieval, materials discovery, and metadata access are uniformly accessible to both human users and autonomous agents through a stable, language-agnostic interface. Additional technical details, including endpoint specifications, authentication mechanisms, and representative workflows, are provided in the Supplementary Information.

%% file: 08_datascribe-agents.tex
\section{DataScribe Agents} \label{datascribe-agents}

\begin{figure}[h]
    \centering
    \includegraphics[width=0.75\linewidth]{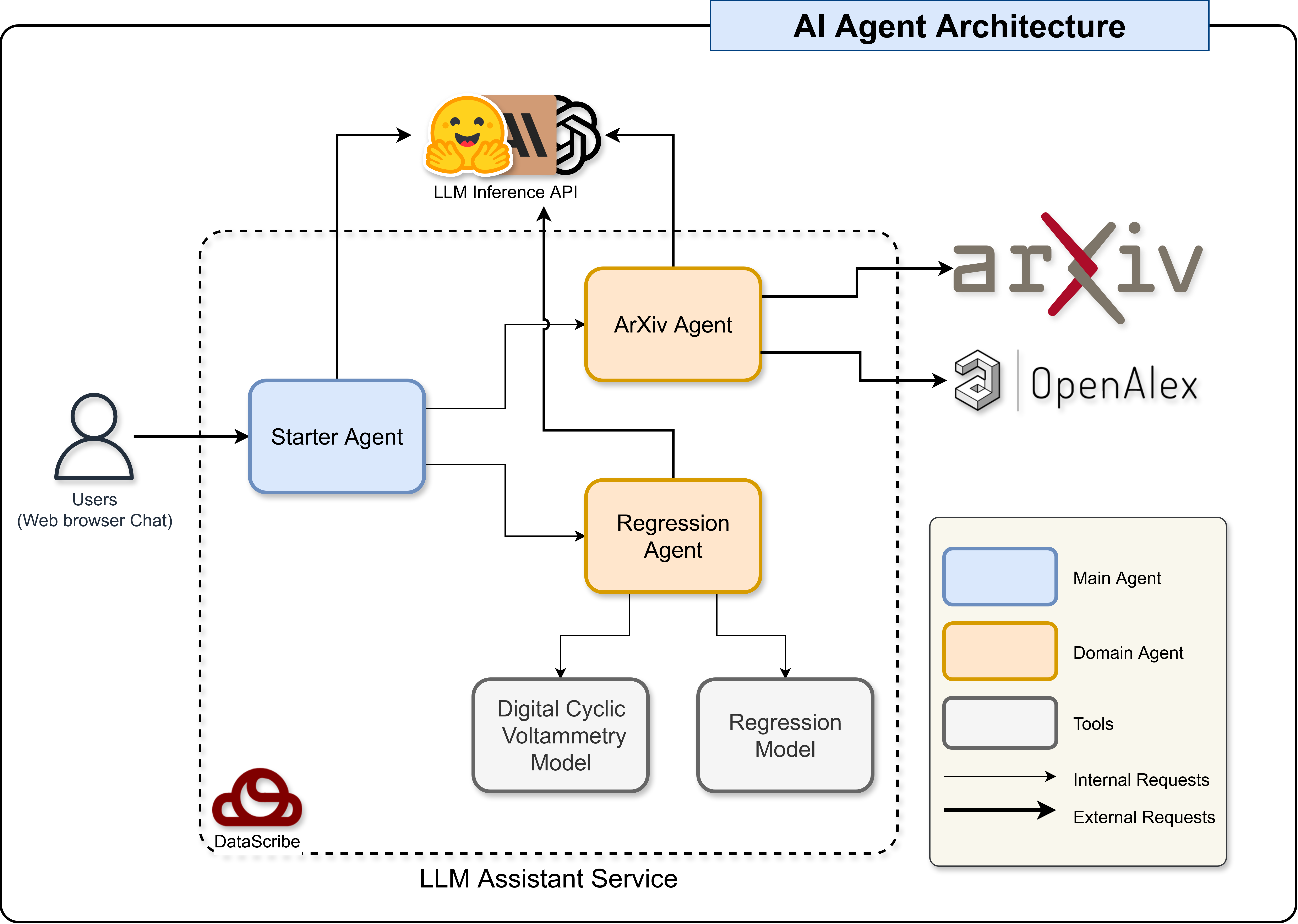}
    \caption{System architecture of the DataScribe LLM Assistant Service illustrating the interaction between internal components and external services. The Starter Agent (orchestrator) coordinates two domain-specific agents: the ArXiv Agent for literature search via ArXiv and OpenAlex APIs, and the Regression Agent for materials property prediction using trained ML models (Voltametry and Regression Models). All agents utilize the HuggingFace Inference API for natural language understanding and generation tasks \cite{huggingface,LangGraph,priem2022openalexfullyopenindexscholarly,arxivAPI}.}
    \label{fig:multi-agent-chatbot}
\end{figure}

DataScribe integrates agentic AI services that connect heterogeneous data with autonomous reasoning and workflow execution. These agents use large language models, adapted for materials science, to interpret user queries, retrieve or synthesize knowledge from internal datasets and external repositories, and coordinate analytical or predictive workflows. The system follows an orchestrator–specialist pattern (Fig.~\ref{fig:multi-agent-chatbot}), where a central agent performs intent recognition and delegates tasks to domain-specific agents. A literature agent conducts rapid mining and summarization of arXiv and OpenAlex records, while a prediction agent interfaces with trained models to provide on-demand property estimates for electrochemical and mechanical systems. Together, these agents create a bridge between DataScribe’s unified data layer and downstream analytics, supporting closed-loop discovery, context-aware decision-making, and digital-twin construction. Detailed routing logic and multi-step decision flows are provided in Supplementary Fig.~\ref{Sfig:llm-agent-decision-process}

%%%%%

%% file: 09_datascribe-application-interface.tex
\section{DataScribe Application Interface}
\label{sec:datascribe-application-interface}

\begin{figure}[h]
    \centering
    \includegraphics[width=0.90\linewidth]{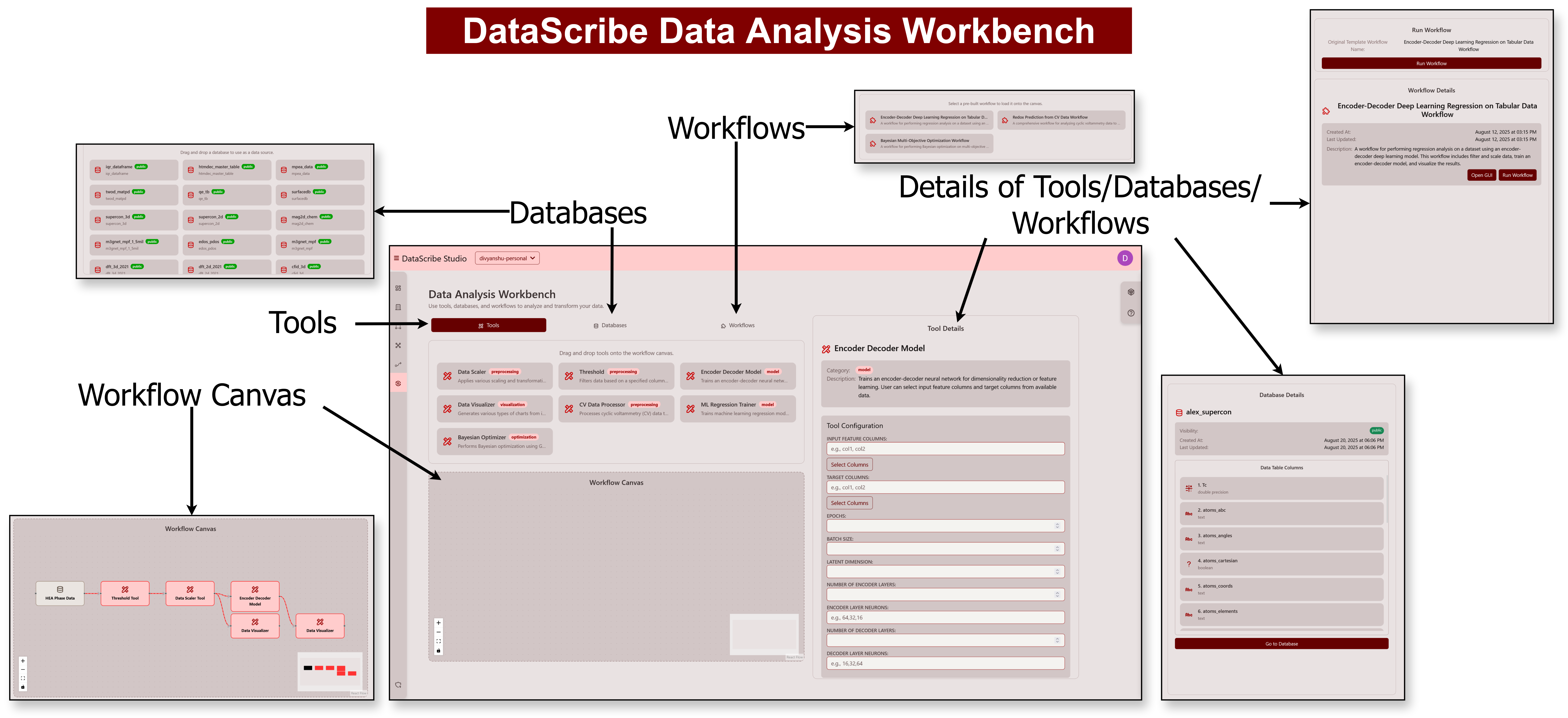}
    \caption{DataScribe Data Analysis Workbench interface showing the organizational structure of Tools, Databases, and Workflows, the React Flow-based visual canvas for drag-and-drop workflow construction, and contextual detail panels for configuration and execution.}
    \label{fig:data-analysis-workbench}
\end{figure}

The DataScribe Application Interface provides an integrated web-based environment for constructing, visualizing, and executing data-driven analytical workflows tailored to materials science applications (Fig.~\ref{fig:data-analysis-workbench}). The interface architecture consists of three primary organizational categories: tools, databases, and workflows, each accessible through a unified navigation panel that facilitates rapid exploration and workflow composition.

The platform's databases section grants direct access to ingested data tables, enabling users to browse available datasets, inspect schema metadata, and preview tabular content before incorporating them into analytical pipelines. The tools section catalogs individual computational modules spanning preprocessing operations, machine learning algorithms, optimization routines, and visualization components, each documented with input–output specifications and usage guidelines. Workflows represent composable sequences of tools and datasets that define end-to-end analytical pipelines, ranging from simple data transformations to complex multi-step machine learning procedures. A contextual details panel on the right displays metadata, configuration parameters, and execution controls for the currently selected tool, database, or workflow.

Central to the interface is a node-based visual canvas implemented using React Flow, which enables intuitive drag-and-drop workflow construction. Users can assemble custom pipelines by positioning dataset nodes, connecting them to preprocessing and modeling tools, and linking outputs to visualization modules, with directed edges representing data flow dependencies. Each node exposes configurable parameters through contextual panels, allowing fine-tuned control over algorithmic behavior, hyperparameters, and output formats. While the graphical workflow editor provides a low-barrier entry point for pipeline design, backend execution capabilities for custom user-defined workflows remain under active development. Current functionality supports visualization and planning of analytical sequences, with full programmatic execution targeted for subsequent platform releases.

%% file: 10_conclusion.tex
\section{Conclusion}

Accelerating materials discovery demands more than data repositories or computational toolkits. It requires an integrated intelligence layer that unifies heterogeneous information, embeds optimization directly into workflows, and closes the loop between hypothesis and validation. In an era where advanced materials underpin semiconductor supply chains, clean energy systems, and defense capabilities, the pace of innovation has become a matter of strategic consequence. DataScribe addresses this imperative by providing an AI-native, ontology-driven platform where experimental, computational, ELN-derived, and literature data converge into machine-actionable knowledge graphs, and where multi-objective Bayesian optimization guides iterative design in real time.

It is important to situate this contribution within the broader laboratory informatics ecosystem. Many capabilities often emphasized in materials platforms—such as unit normalization, schema mapping, FAIR metadata capture, and structured workflow or traveler forms—are now standard across established ELNs and LIMS. Researcher-facing ELNs like Benchling, LabArchives, and ChemOS, and operations-facing LIMS such as LabWare, STARLIMS, and Thermo Fisher SampleManager already offer schema-on-write data models, controlled vocabularies, provenance tracking, and integration with laboratory instruments. Materials-focused systems including Citrination, Materials Commons, AiiDA, and MAPI provide comparable functionality for metadata harmonization and cross-format ingestion. These features have become broadly commoditized, and therefore do not, on their own, constitute innovation.

In contrast, DataScribe provides an AI-native data substrate that integrates a hybrid relational and graph semantic layer engineered specifically for multi-objective optimization, agentic workflows, and rapid model-experiment iteration. Incoming data streams are contextualized through dynamically extensible ontologies that capture provenance, hierarchical relationships, and dependencies among CALPHAD grids, CV curves, DFT calculations, microstructure images, ELN entries, and process logs. This semantic substrate enables autonomous agents to perform contextual ingestion, metadata enrichment, schema inference, and linkage of heterogeneous datasets into a unified knowledge representation that feeds directly into Bayesian optimization, uncertainty-aware modeling, and digital twin construction.

The platform’s capabilities were demonstrated through pre-built analytical workflows: deep learning regression on tabular materials data, prediction of electrochemical responses from cyclic voltammetry, and uncertainty-guided multi-objective optimization across competing performance metrics. A RESTful API and Python client support programmatic integration, while a multi-agent assistant provides natural-language access to datasets and literature synthesis. Underlying these tools is a governed workflow that spans organization management, ontology and schema design, metadata-rich ingestion from ELNs and computational pipelines, and collaborative analysis with full provenance.

Future versions will allow execution of user-defined workflows through the graphical interface, connect directly to laboratory instruments and HPC schedulers, and expand to microscopy, spectroscopy, and operando data streams. Planned analytical extensions include hybrid digital twins that couple thermodynamic models and phase-field simulations with learned surrogates, along with policy-aware optimization that incorporates critical-mineral exposure, domestic sourcing likelihood, and embodied carbon constraints.

By operationalizing intelligence at the application layer, DataScribe offers a pathway to compress materials innovation cycles from decades to months. For the research community, it transforms fragmented datasets into interoperable, FAIR-aligned knowledge. For industry, it accelerates qualification timelines and reduces the risk associated with adopting new materials. For policymakers, it provides an infrastructure that aligns materials R and D with energy security, defense readiness, and resilient supply chains. As the field moves toward networks of autonomous self-driving laboratories, DataScribe establishes the data backbone and decision-support architecture needed to support that future and sustain U.S. leadership in materials innovation.

%% file: 11_methods.tex
\section{Methods}

DataScribe is structured around an ontology-backed intelligence stack through five methodological pillars: (i) a cloud-native, microservices architecture that supports scalable analysis and workflow execution; (ii) structured traveler forms that enforce metadata-rich data capture aligned with electronic laboratory notebook (ELN) conventions; (iii) schema templates that encode domain knowledge and enable interoperable, cross-group data organization; (iv) a RESTful API gateway providing unified programmatic access to internal datasets and external materials databases; and (v) multi-agent orchestration logic that automates contextual reasoning, model construction, and workflow execution. These components integrate experimental and computational activities within a closed-loop discovery system that supports real-time optimization, uncertainty-guided sampling, and decision-making under explicit constraints.

\subsection{Deployment Details: Cloud-Native Infrastructure and Microservices Architecture}

DataScribe is deployed as a cloud-native application on DigitalOcean~\cite{digitalocean} and orchestrated using MicroK8s Kubernetes~\cite{microk8s}. Seven containerized microservices form the backbone of the platform, each communicating through REST APIs~\cite{richardson2007restful}. NGINX performs reverse-proxy routing and load balancing~\cite{nginx}, enabling scalable execution under fluctuating computational demand typical of high-throughput discovery campaigns. Authentication relies on Google OAuth 2.0~\cite{hardt2012oauth}, enabling secure federation across institutions. The microservices include: the Backend API Service (system logic and orchestration), Frontend (React.js~\cite{react}), Workflow GUI (node-based pipeline construction), Data Table Service (high-performance tabular operations), Data Analysis Service (FastAPI~\cite{fastapi}-based ML and analysis tools), LLM Assistant Service (agentic reasoning via Hugging Face~\cite{huggingface}), and an Authentication Service implementing federated identity. This architecture supports integration with cloud storage (Google Drive API~\cite{google-drive-api}), materials databases (MP~\cite{jain2013commentary}, AFLOW~\cite{curtarolo2012aflow}, OQMD~\cite{OQMD_API_1}), and future autonomous laboratory instrumentation. Kubernetes~\cite{kubernetes} provides automatic scaling, failure isolation, and continuous execution required for uninterrupted optimization workflows.

\subsection{Traveler Forms: Structured ELN-Aligned Data Capture and Semantic Enforcement}

Traveler forms enable metadata-rich ingestion by aligning experimental, computational, and electronic laboratory notebook (ELN), derived records with predefined semantic structures~\cite{wilkinson2016fair}. Rather than treating metadata as a \emph{post hoc} annotation step, as is common in traditional ELNs, DataScribe captures provenance, units, controlled vocabularies, and domain semantics at the point of data upload. Forms are defined either through a graphical interface with configurable field types, validation logic, conditional branching, and explicit ontological mappings or uploading user-specific JSON schema. Each form is associated with specific folders within the organizational hierarchy, such that uploading files to these locations triggers the corresponding form and enforces structured ingestion of compositions, processing conditions, measurement parameters, uncertainty estimates, and instrument metadata. As an illustrative example, an alloy composition traveler form developed for a high-entropy alloy (HEA) consortium enforces constraints on element symbols, atomic fractions, synthesis parameters, required metadata fields, and associated data uploads (e.g., microscopy images, differential scanning calorimetry data). Submitted records are validated, assigned persistent UUIDs, and inserted into governed database tables with full provenance tracking~\cite{mons2018data}. Ontological mappings ensure that fields such as scan rate, microstructure image, or composition formula remain interpretable to downstream agents irrespective of group-specific naming conventions, enabling consistent reasoning across multimodal datasets and automated workflows.

\subsection{Schema Templates: Encoding Domain Knowledge and Enabling Interoperability}

Schema templates define the structural and semantic representation of datasets across research groups. Rather than imposing a monolithic data model, DataScribe supports three schema archetypes, including (1) research schemas emphasizing iteration tracking and reproducibility, (2) industry QMS schemas enforcing compliance and traceability, and (3) agentic schemas that formalize interfaces between data ingestion, model training, and experiment recommendation. Each schema specifies fields, data types, units, nullability constraints, and ontological annotations~\cite{franconi2011quelo}. Ontological tags map columns to domain concepts (e.g., \texttt{ThermodynamicProperty}, \texttt{CrystalStructureProperty}), enabling agents to associate composition data, XRD measurements, and density functional theory predictions without manual harmonization. These templates instantiate PostgreSQL tables with enforced relational integrity and JSON-based metadata extensions for flexibility. Schema reuse supports multi-site scaling, as the same template can underlie multiple independent databases across teams, reducing metadata fragmentation. Because schema semantics are encoded through ontologies rather than rigid table structures, DataScribe accommodates evolving scientific workflows, particularly in self-driving laboratory (SDL) settings, without disruptive schema migrations.

\subsection{API Gateway: Unified Access Across Internal Data and External Materials Databases}

The API gateway provides RESTful access for ingestion, querying, metadata inspection, and materials search. Functionality spans: (i) metadata discovery; (ii) structured data retrieval with JSON-based filters supporting numerical, text, and logical operations; and (iii) property search across external databases (MP, AFLOW, OQMD) normalized into consistent formats. The \emph{datascribe\_api} Python client~\cite{datascribe_api_2025} integrates seamlessly with NumPy, pandas, scikit-learn, and PyTorch~\cite{NEURIPS2019_9015}, enabling researchers and agents to retrieve data, train models, and submit optimization recommendations without manual formatting. OAuth tokens enforce fine-grained access control. Unauthorized requests are blocked and logged, ensuring that agentic workflows operate within the same governance boundaries as human researchers~\cite{foster2011globus}. Ontological query mediation allows domain-level queries (e.g., “find all cubic materials with band gap > 2.5 eV”) to run across multiple materials databases without requiring provider-specific syntax.

\subsection{Agent Orchestration Logic: Automating Contextual Reasoning and MOBO Execution}

Agentic workflows are implemented using the LangGraph framework~\cite{LangGraph}, enabling stateful, interpretable reasoning pipelines. The orchestrator agent parses user intent and delegates tasks to specialized agents (literature retrieval, regression inference, materials search, or optimization). Each decision is expressed in structured JSON with validation and fallback mechanisms to prevent brittle or hallucinated outputs~\cite{newman2015building}. This architecture powers the closed-loop propose–measure–learn cycle. Agents retrieve research data (either organization or literature data) through the API, train surrogate models (Gaussian processes~\cite{rasmussen2003gaussian}, neural encoder-decoder networks), evaluate acquisition functions, and return structured experiment recommendations. Explanations (SHAP attribution, uncertainty overlays, Pareto visualizations) maintain human-in-the-loop oversight. Audit trails capture all actions, ensuring traceability and compliance for industrial or multi-institution campaigns.

%\subsection{Integrated Operation: Ontology-Driven Data Fusion and Real-Time Optimization}

\begin{figure}[!ht]
    \centering
    \includegraphics[width=0.90\linewidth]{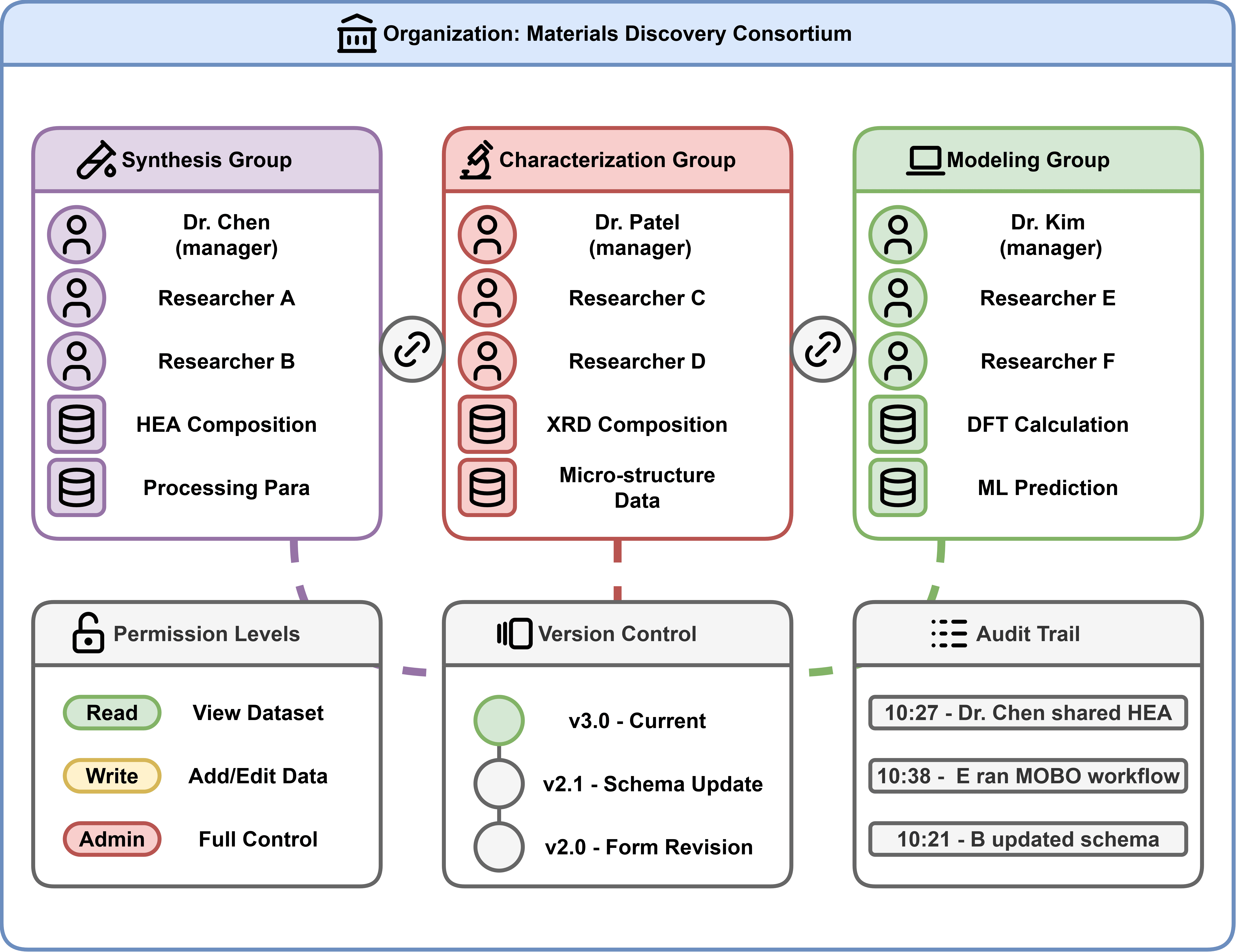}
    \caption{Collaboration and governance framework showing cross-group data sharing within a multi-team organization. Research groups maintain independent datasets while selectively sharing data through role-based permissions configured in Organization Management. The platform tracks all modifications through comprehensive audit trails and preserves version history for reproducibility and compliance.}
    \label{fig:collaboration-infographic}
\end{figure}

As shown in Figure~\ref{fig:collaboration-infographic}, DataScribe supports versioning, audit trails, controlled collaboration, and cross-institution interoperability, critical features for deploying agentic materials discovery workflows at scale. The previous five components integrate into an application-layer intelligence stack. Traveler forms guarantee that ELN-style records enter the system with semantic consistency. Schema templates provide interoperable structure across teams. The API gateway connects internal data with external knowledge bases. The cloud-native architecture ensures scalable execution. Agent orchestration completes the closed-loop, enabling autonomous optimization within governed, shareable, multi-team environments. This integration operationalizes the A–Z materials acceleration framework shown in Table~\ref{tab:SDL_AZ_Framework} by unifying data infrastructure, learning pipelines, active learning feedback, and optimization under a single ontology-driven substrate. 

%% file: 12_codeAvailability.tex
\section*{Author contributions}

Divyanshu Singha contributed to frontend and backend platform development, architectural conceptualization, integration of machine-learning and Bayesian optimization workflows, and preparation of figures and manuscript content. Do\u{g}uhan Sar\i{}t\"urk led the design and development of the DataScribe API and contributed to platform integration. Cameron Lea contributed to the development of multi-agent workflows and supporting system components. Md Shafiqul Islam contributed to the development and validation of the multi-objective, multi-fidelity Bayesian optimization workflow. Raymundo Arroyave provided scientific oversight, strategic guidance, and funding acquisition. Vahid Attari conceived and directed the project, designed the platform architecture and scientific framework, supervised software and methodological development, provided conceptual guidance on materials design and optimization frameworks, and led manuscript writing and revision. 

\section*{Conflicts of interest}

There are no conflicts to declare.

\section*{Data and Code Availability}

The DataScribe platform is accessible at \url{https://datascribe.cloud}. The \textit{datascribe\_api} Python client is publicly available on GitHub (\url{https://github.com/DataScribe-Cloud/datascribe_api}) and installable via PyPI (\texttt{pip install datascribe-api}). API documentation and usage tutorials are provided through the platform's documentation portal at \url{https://documentation.datascribe.cloud}.

\newpage